\documentclass[twocolumn]{IEEEtran}
\usepackage{graphicx}
\usepackage{cite}
\usepackage{picinpar}
\usepackage{amsmath}
\usepackage{amsfonts}
\usepackage{stfloats}
\usepackage{url}
\usepackage{flushend}
\usepackage[latin1]{inputenc}
\usepackage{colortbl}
\usepackage{soul}
\usepackage{multirow}
\usepackage{pifont}
\usepackage{color}
\usepackage{alltt}
\usepackage{enumerate}
\usepackage{siunitx}
\usepackage{breakurl}
\usepackage{epstopdf}
\usepackage{pbox}
\usepackage{mathrsfs}
\usepackage[]{caption2} 

\usepackage{algorithm}
\usepackage{algorithmic}

\usepackage{booktabs}
\usepackage{bigstrut}
\usepackage{multirow}
\usepackage{rotating}
\usepackage[table,xcdraw]{xcolor}

\usepackage{subfigure}

\makeatletter
\long\def\@makecaption#1#2{\ifx\@captype\@IEEEtablestring%
	\footnotesize\begin{center}{\normalfont\footnotesize #1}\\
		{\normalfont\footnotesize\scshape #2}\end{center}%
	\@IEEEtablecaptionsepspace
	\else
	\@IEEEfigurecaptionsepspace
	\setbox\@tempboxa\hbox{\normalfont\footnotesize {#1.}~~ #2}%
	\ifdim \wd\@tempboxa >\hsize%
	\setbox\@tempboxa\hbox{\normalfont\footnotesize {#1.}~~ }%
	\parbox[t]{\hsize}{\normalfont\footnotesize \noindent\unhbox\@tempboxa#2}%
	\else
	\hbox to\hsize{\normalfont\footnotesize\hfil\box\@tempboxa\hfil}\fi\fi}
\makeatother

\begin{document}
\title{Multi-Task Learning Enhanced Single Image De-Raining}

\author{
	\vskip 1em
	Yulong Fan, Rong Chen*, Bo Li
	
	\thanks{
		
		
This work is supported by the National Natural Science Foundation of China (No. 61672122, No. 61402070, No.61602077), the Natural Science Foundation of Liaoning Province of China (No. 20170540097No. 2015020023), and the Fundamental Research Funds for the Central Universities (No. 3132016348), Next-Generation Internet Innovation Project of CERNET (No.NGII20181205). (\emph{Corresponding author: Rong Chen}.)

Y. Fan, R. Chen and B. Li are with the College of Information Science and Technology, Dalian Maritime University, Dalian, 116026, China (e-mail: sumihui@hotmail.com; rchen@dlmu.edu.cn; delate@126.com).

		}
}

\maketitle
	
\begin{abstract}
Rain removal in images is an important task in computer vision filed and attracting attentions of more and more people. In this paper, we address a non-trivial issue of removing visual effect of rain streak from a single image. Differing from existing work, our method combines various semantic constraint task in a proposed multi-task regression model for rain removal. These tasks reinforce the model's capabilities from the content, edge-aware, and local texture similarity respectively. To further improve the performance of multi-task learning, we also present two simple but powerful dynamic weighting algorithms. The proposed multi-task enhanced network (MENET) is a powerful convolutional neural network based on U-Net for rain removal research, with a specific focus on utilize multiple tasks constraints and exploit the synergy among them to facilitate the model's rain removal capacity. It is noteworthy that the adaptive weighting scheme has further resulted in improved network capability. We conduct several experiments on synthetic and real rain images, and achieve superior rain removal performance over several selected state-of-the-art (SOTA) approaches. The overall effect of our method is impressive, even in the decomposition of heavy rain and rain streak accumulation. The source code and some results will be released at: https://github.com/SumiHui/MENET.
\end{abstract}


{}

\definecolor{limegreen}{rgb}{0.2, 0.8, 0.2}
\definecolor{forestgreen}{rgb}{0.13, 0.55, 0.13}
\definecolor{greenhtml}{rgb}{0.0, 0.5, 0.0}

\section{Introduction}

\IEEEPARstart{I}{mages} of outdoor scenes are sometimes accompanied by rain streak. With the interference of rain, the content and color of images are often drastically altered or degraded. The undesirable corruption may have negative impact on most outdoor vision systems, such as surveillance and autonomous navigation.

Typically, research progress depends on better assumptions and priors about low-level features such as HoG, patch-based priors, and sparseness. But they may fail miserably if the defined features are interfered with environmental factors such as atmospheric veils and strong light. Note that rain streak may appear anywhere with different intensity, brightness and directions. There is a complex interaction of rain streak and interfering factors. So robust features have remained stubbornly difficult to develop.  As a result, some methods \cite{kang_id,lp,dsc} have difficulty in coping with these adverse factors in real rain images. This issue is even more salient for many existing methods if they work on local image patches or a limited spatial range without sufficient spatial contextual information.

\begin{figure}[t]\centering  
	\includegraphics[width=8.5cm]{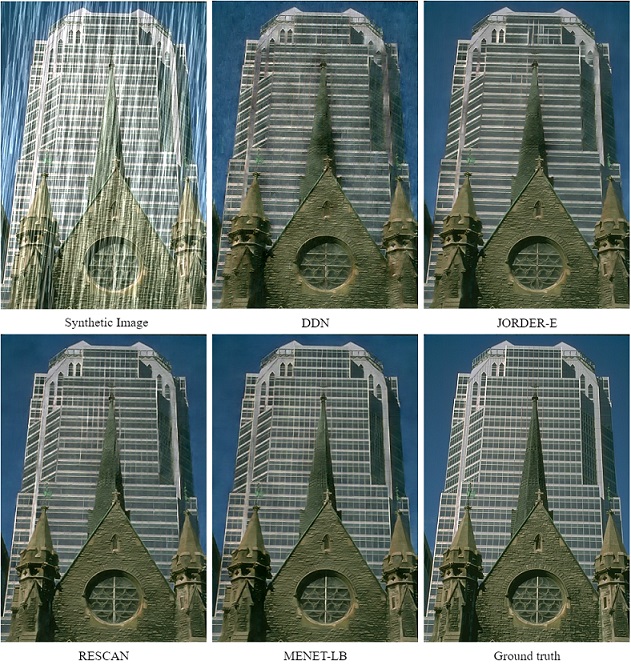}	
    \caption{Comparing the state of the art by DDN \cite{ddn}, JORDER-E \cite{jorder_e} and RESCAN \cite{rescan} with the sharper, perceptually more plausible result produced by MENET-LB on an image from Rain100H.}
    \label{home_show}
\end{figure}

One appealing solution is convolutional neural networks (CNNs) that can provide non-linear transformations of the extracted features with spatial contextual information. With hundreds of feature maps detected automatically in multiple layers, CNNs have demonstrated a massive impact on performance of computer vision such as object detection \cite{fast_rcnn,faster_rcnn} and semantic segmentation \cite{he_instance_aware} because they perform non-linear transformations of the extracted features with spatial contextual information. Also a few CNN methods \cite{ddn,rescan,jorder,scale_aware_derain} have been developed for rain streak removal and empirically tested to work in a limited range of settings. However, some existing methods are slightly inadequate in terms of edge sharpening and texture detail recovery, see Fig. 1. Most recently a widely used machine learning technique---Multi-task learning (MTL) \cite{multitask_learning} has been incorporated into variants of CNNs (e.g. \cite{fast_rcnn,faster_rcnn,he_instance_aware,multi_task_face_recog}), and significant progresses have achieved on object detection (e.g. \cite{faster_rcnn,fpn}) and semantic segmentation (e.g. \cite{he_instance_aware,fcn}). Such studies are unique in that they typically exploit the synergy among the tasks that operate on the extracted features and boost up their individual performances. In doing so, some techniques on object detection treat multiple tasks as outputs (e.g. object type and location etc.), while other techniques use auxiliary tasks to train one major task instead of maximizing the performance of all tasks. To deal with the de-raining problem, the researchers \cite{jorder,jorder_e,did-mdn} are also considering increasing constraints related to image quality descriptions. However, few papers have looked at developing better mechanisms for MTL in CNN-based single image de-raining.

The aim of the present paper is to validate multi-task enhanced CNN specially designed for rain removal research, with a specific focus on how to learn more complex interactions between tasks to ensure the de-raining performance. We note that CNN-based rain removal methods often operate on the extracted features at two different semantic levels: perceptual level \cite{perceptual_loss} and pixel level. The former is used to restore the image background based on differences between image features extracted from transform domain, while the latter is used to separate the rain-free image based on differences between pixel information. In our framework, we designed two semantic level task constraints and one pixel level task constraints. And two dynamic weighting algorithms are proposed to exploit the synergy among them to build a powerful de-raining system. 

Our contributions can be summarized as follows:

\begin{enumerate}[(1)]
\item We introduce de-subpixel and channel attention to get a powerful model for the deep image representation. And also show that our de-raining network are powerful enough to be useful for exploit the synergy among multi-tasks, although we need to further explore how to design appreciate task constraints and handle multiple tasks better for this goal.

\item The main difference in constructing a loss network is that we use specific weights instead of pre-trained weights. We employ a set of isotropic image gradient operators as the filter kernels to construct a perceptual level loss, it forces the de-raining network to perform edge detection as a joint task in image reconstruction and facilitates the edge preservation for de-rained results.

\item Based on local texture similarity, we proposed a novel loss function to remove the rain streaks from the photograph. This constraints can not only successfully removes most rain streaks but also preserving most original image details. The experimental results have shown that proposed has good quality in rain removal. 

\item Another change is that we apply an adaptive approach to build a powerful network to stabilize the visualization quality. We make progress on the challenges of weight problems in multi-task learning by designing an appropriate dynamic task weighting framework, potentially applicable to other image tasks. Designing an appropriate objective function that handles the uncertainty of the de-raining problem.

\end{enumerate}

The rest of the paper is organized as follows. In section 2, we review parallel, recent work that have state-of-the art results in de-raining task. In section 3, we present details of the proposed multi-task regression model and network architecture. In section 4, comprehensive experiments are presented. At last, we conclude our work in the section 5.

\section{Related work}

In this section, we give a brief review on the most related work on rainy image processing. To improve the visibility of images, there have been a few approaches \cite{kang_id, dsc, lp, ddn, did-mdn, jorder,jorder_e,rescan} proposed to remove rain streak from individual image.

Kang et al. \cite{kang_id} proposed a rain removal framework (ID) based on a morphological component analysis. They firstly decomposed an image into the low-frequency and high-frequency parts using a bilateral filter. Then an attempt was made to separate the high-frequency part by using the dictionary learning algorithm with HoG \cite{hog} features. However, the overall framework in \cite{kang_id} is complex and performs poorly on images with complex structures.

Luo et al. \cite{dsc} proposed a dictionary learning based algorithm (DSC) to recover a clean image from a rainy image. Their method sparsely approximated the patches of image and rain layers by very high discriminative codes over a dictionary learning, but it is not easy to learn such a dictionary with strong mutual exclusivity property.

Besides dictionary learning based algorithm, an adaptive nonlocal means filter algorithm (ANLMF) \cite{nonlocal_means_filter} was proposed to remove rain from a single image by assuming that rain streak has elongated elliptical shapes. They used a kernel regression method to determine the locations of the rain streak regions, and then restore the rain streak regions using the nonlocal means filter. Experiments of ANLMF \cite{nonlocal_means_filter} demonstrated better results than that of ID \cite{kang_id}.

Recently, Yu et al. \cite{lp} proposed a method (LP) based on Gaussian mixture models in which patch-based priors were used for both a clean layer and a rain layer. The authors showed how multiple orientations and scales of rain streak can be accounted for by such pre-trained Gaussian mixture models. Despite that it works for some cases, some relevant spatial information about rain streak may be lost in general.

In recent research, to handle nonlinear input-output relationship for the rain detection and removal, deep learning based methods  \cite{ddn} and \cite{jorder} have recently been emerged as an appealing solution.

Inspired by the deep residual network (ResNet) \cite{he_resnet}, Fu et al. \cite{ddn} proposed a deep detail network (DDN) architecture for removing rain streak from individual image. The basic idea behind ID \cite{kang_id}, DSC \cite{dsc} and DDN \cite{ddn} is to exploit the highly frequent information such as rain streak and object structures, remained only in the high-frequency detail layer, yet spare. But they differ from the treatment of such information; ID \cite{kang_id} and DSC \cite{dsc} are not connectionist, whereas DDN \cite{ddn} is a neural-network based approach that exploits the high frequency detail layer more thoroughly. Based on residual learning experience, shrinking solution space can improve the network performance. Putting them together, they introduced guided filtering \cite{he_guided_filter} as a low-pass filter to split the rainy image into base and detail layers, and then trained a novel ResNet on the high frequency detail layer. DDN showed that combining the high frequency detail layer content of an image and regressing on the negative residual information have benefits for de-rain performance.

It is worth noting that JORDER \cite{jorder} is a method which combined rain mask, rain streak, and background. Based on a region-dependent model, Yang W. et al. \cite{jorder} introduced Multi-task Network Cascades (MTNC) \cite{he_instance_aware} into a recurrent rain detection and removal neural network. The MTNC features mask layer to locate objects, while JORDER uses it to identify regions with rains. JORDER has been proved very effective for removing rains and rain accumulation, and also for removing mist surrounding rain streak if combining with haze removal method.In a follow-up study, the authors further proposed an enhanced version??JORDER-E \cite{jorder_e}, in which an extra detail preserving step is introduced. 

Following Yang et al. \cite{jorder} and Fu et al. \cite{ddn}, many methods based on deep CNNs \cite{rescan,zhanghe_cgn_derain,did-mdn,non_local_derain} are proposed, by employing more advanced network architectures or injecting new rain-related priors. Instead of relying on image decomposition framework like \cite{ddn}, Zhang et al. \cite{zhanghe_cgn_derain} proposed a conditional generative adversarial networks (GAN) for single image de-raining which incorporated quantitative, visual, and discriminative performance into objective function. Since a single network may not learn all patterns in training samples, the authors \cite{did-mdn} further presented a density-aware image de-raining method using a multi-stream dense network (DID-MDN). By integrating a residual-aware classifier process, DID-MDN can adaptively determine the rain-density information (heavy/medium/light).

Similar to Yang et al. \cite{jorder}, Li et al. \cite{rescan} proposed a recurrent squeeze-and-excitation based context aggregation network (RESCAN) for single image rain removal, where squeeze-and-excitation block assigned different alpha-values to various rain streak layers and context aggregation network acquired large receptive field and better fit the rain removal task.

Unlike existing deep learning methods, which typically uses only local information on the previous layer at each layer, Li et al. \cite{non_local_derain} proposed a non-locally enhanced encoder-decoder network to efficiently learn increasingly abstract feature representation for more accurate rain streaks and then finely preserve image details, but the consumption of hardware resources and computations is too large.

\section{Our method}

In this section, we will present the proposed de-rain system which features a multi-task regression model that includes one reconstruction task for rain removal and multiple auxiliary tasks to further optimize the model predictions. Also we will show the simple yet powerful CNN architecture for de-rain task. We predict results that are closer to the ground truth through multiple task constraints. In addition, we re-weight the loss at training time to emphasize the balance of the task?s impact.

\subsection{Problem formulation}

Like the commonly used rain model \cite{lp,dsc}, we can characterize the observed rainy image ${\mathbf{O}} \in {\mathbb{R}^{C \times H \times W}}$ as a linear superimposition of the desired background layer and the rain streak layer:
\begin{equation}
{\mathbf{O}} = {\mathbf{B}} + {\mathbf{R}},
\end{equation}
such that ${\mathbf{B}} \in {\mathbb{R}^{C \times H \times W}}$ and ${\mathbf{R}} \in {\mathbb{R}^{C \times H \times W}}$ are the desired background layer and the rain streak layer, respectively, where $H \times W$ is the spatial dimension and $C$ represents the number of channels.

\begin{figure}[t]
\begin{center}
\includegraphics[width = 8.5cm]{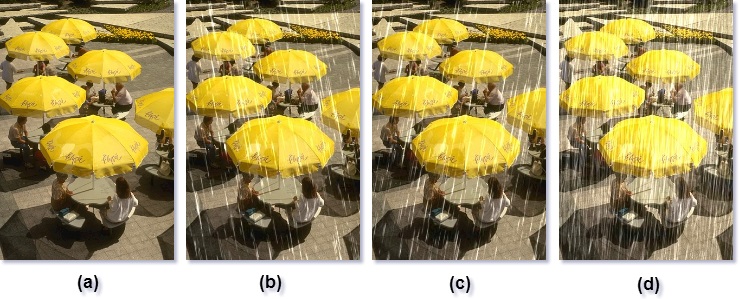}	
\caption{Schematic of ill-posed. (a) shows the rain-free image, (b-d) are rainy images with different conditions. The rain streak in (d) is too dense to clear the connection with (a).}
\label{fig:ill_posed}
\end{center}
\end{figure}

We can assume that the desired background layer ${\mathbf{B}}$  is independent of the rain streak layer ${\mathbf{R}}$. Albeit simple, this formulation can appropriately represent a general peculiarity of rain streak. So the ultimate goal of our task is to separate the rain-free background layer ${\mathbf{B}}$  and the rain layer ${\mathbf{R}}$  from a given rainy image ${\mathbf{O}}$. Obviously, this is an ill-posed problem as shown in Fig. \ref{fig:ill_posed}. In many ways, rain streaks superimpose their visual effects onto the same background image, depending on their sizes and velocities and angles. The causal link may become more ambiguous if the rain streak is denser, because fine details of the rainy image may have little or no hints on its rain-free version.

\subsection{Loss function}

To address this ambiguity well, we propose a multi-task regression model to advance the de-raining network. We consider features at two semantic levels, on which we design a series of tasks. The main task is to reconstruct background layer from rainy image, while a set of auxiliary tasks incorporate other favorable elements to facilitate the learning of model. The total loss can be expressed as
\begin{equation}
L_{\mathrm{total}}=\sum_{i} w_{i} L_{i}
\label{fomula:total_loss}
\end{equation}

Where $L_{i}$ denotes the losses of a task, $w_{i}$ is non-negative coefficients to determine the importance of the corresponding task, and $i \in\{p,e,t\}$. The Loss Function is used to measure the discrepancy between the model's predictable value and the true value from all aspects. Our goal is to minimize the loss function and bring the model's predictions as close as possible to the ground truth.

\paragraph{\textbf{Reconstruction loss}}
A per-pixel loss function is the easiest way to measure the difference between expected result $y$ and actual output $\hat{y}$ of neural network. The popular empirical training criteria is mean square error (MSE) which produces normalized Euclidean distance as follows:
\begin{equation}
{\ell _{pixel}}({\mathbf{y}},{\mathbf{\hat y}}) = \frac{{\left\| {{\mathbf{y}} - {\mathbf{\hat y}}} \right\|_F^2}}{{CHW}},
\end{equation}
where ${\left\| \cdot \right\|_F}$  represents the Frobenius norm, we denote predictions with a $\hat{\bullet}$ symbol and ground truth without, $ {\mathbf{\hat y}}$  is the output image and $ {\mathbf{y}}$  is the target, and both have shape $C \times H \times W$. Based on Eq. (2), the the reconstruction loss of our model can be characterized as:
\begin{equation}
{L_p}{\text{ = }}{\ell _{pixel}}({\mathbf{B}},{\mathbf{b}}),
\label{Lp}
\end{equation}
where ${\mathbf{b}}$  represents the approximation to the background.

Pixel-wise L2 distance can be used as the single loss function, and all the local areas are processed equally. The Downside is without considering image regular structures. To solve this issue, we further introduced two novel loss functions (named as edge-aware loss and texture matching loss) by constraining the edge distortion and texture similarity. This is because deep learning allows us to easily incorporate problem specific constraints (or priors if we prefer) directly into the model to reduce prediction errors.

\paragraph{\textbf{edge-aware loss}}
It guarantees the similarity between the predicted image and the target image in edge feature expression. It aims to produce images with sharp edge. Similar to perceptual loss \cite{perceptual_loss}, we also let edge-aware loss defined as the Euclidean distance between multiple feature maps of a specific layer in a network:
\begin{equation}
\ell_{\text {edge-aware}}^{\phi}(y, \hat{y})=\frac{\|\phi(y)-\phi(\hat{y})\|_{F}^{2}}{C H W}
\label{edge-aware-loss}
\end{equation}
where ${\phi}$ represents a neural network with fixed model parameters. In perceptual loss, this network is usually a pre-trained VGG16/19 network on the ImageNet classification dataset.

As we know, edge is one of the most important aspect of the human visual assessment, and human observers are always sensitive to edge distortions. However, traditional reconstruction methods treated all the pixels equally and failed to recover the sharp edges in an effective way. To encode the edge texture information and force the de-raining network to perform edge detection in rainy image reconstruction, the loss network ${\phi}$ we utilize is a shallow feed-forward neural network, whose kernels are determined by the Sobel filters. In this case, we also refer to perceptual loss as edge-aware loss. By integrating edge texture information to the training of our backbone network, we can let the model pay more attention on the edge details. It is worth noting that edge-aware loss also has the ability to constrain area smoothing, because edge-aware loss magnifies this difference if the model predicts the original smoothed region to be unsmooth.

\par Based on Eq.(5), the perceptual loss of our model can be computed as:

\begin{equation}
L_{e}=\ell_{\text {edge-aware}}^{\phi}(\mathbf{B}, \mathbf{b})
\label{Le}
\end{equation}

\begin{figure*}[t]	
	\centering
	\includegraphics[width=1\textwidth]{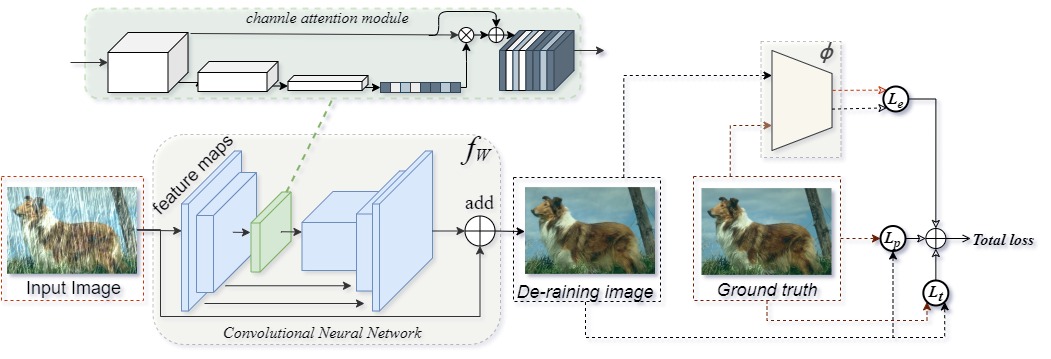}		
	\caption{Overview the pipeline of proposed MENET system. The proposed MENET precedes as follows: First, it down samples and encodes the input into a feature map, extracts local features via a series of convolution layer. Then it up-samples the result to produce the full-resolution multi-channel residual information, and takes it to recover the full-resolution de-raining image. In order to improve the prediction accuracy, it also incorporate the loss network to provide higher performance. Note that loss network is not trainable. We train the end-to-end network to learn $R$ from image pairs $\{O, B\}$ with three loss components $\left\{L_{p}, L_{e}, L_{t}\right\}$. Arrows represent the main data flow of feed-forward. }
	\label{pipeline_of_our_system}
\end{figure*}

\paragraph{\textbf{Texture matching loss}}
 Johnson et al. \cite{perceptual_loss} and Gatys et al. \cite{gatys2016_style_transfer,gatys2015texture} demonstrate how style reconstruction loss can be used to create high quality textures. In image style transfers, Gram Matrix loss is able to migrate textures. So we are motivated to use it on image recovery tasks for the purpose of texture-preserving. To ensure faithful texture generation and the overall perceptual quality of the images, we specifically define the texture matching loss which extends style reconstruction loss. By exploiting the local self-similarity of the image to emphasize the texture influence. Formally, the texture matching loss is the squared Frobenius norm of the difference between the Gram matrices of the output patches and target patches:

\begin{equation}
\ell_{\text {texture}}(y, \hat{y})=\left\|G\left((y)_{p}\right)-G\left((\hat{y})_{p}\right)\right\|_{2}^{2}
\end{equation}

With Gram matrix  $G(F)=F F^{T} \in \mathbb{R}^{n \times n}$. Where $(\bullet)_{p} \in \mathbb{R}^{C K^{2} \times M}$  denotes patch operation with patch size $K \times K$  are used, and $M=(H W) / K^{2}$ . In this paper, we set $K=4$ . Based on Eq.(7), our texture matching loss expressed as:

\begin{equation}
L_{t}=\ell_{\text {texture}}(\mathbf{B}, \mathbf{b})
\label{Lt}
\end{equation}

We compute the texture matching loss  $\ell_{\text {texture}}$  patch-wise during training to enforce locally self-similar textures between $b$ and $B$. The network therefore learns to produce images that have the same local textures as the rain-free images during training. For further details on the implementation, we refer the reader to the code.

\subsection{Network architecture}

As shown in Fig. \ref{pipeline_of_our_system}, our network architecture is composed of two modules: a de-rain network  ${f_{\mathbf{w}}}$ and an edge-aware loss network ${\phi}$. The de-raining network is built with 8 residual blocks, which transforms input rainy image ${\mathbf{O}}$  into output image ${\mathbf{b}}$, characterized by the mapping ${f_{\mathbf{w}}}:{\mathbf{O}} \to \mathbf{b}$.Skip connection \cite{he_resnet,he_identity_map} is used to ease the training of de-raining network. For loss network, it takes $B$  and $b$  as inputs and outputs their feature maps. Then Eq.(6) is used to figure out the distance between $\phi ({\mathbf{B}})$  and $\phi ({\mathbf{b}})$. Specifically, our framework does not ship with a pretrained VGG network, instead we use a shallow feed forward network.

\begin{figure}[ht]
	\centering  
	\includegraphics[width=0.3\textwidth]{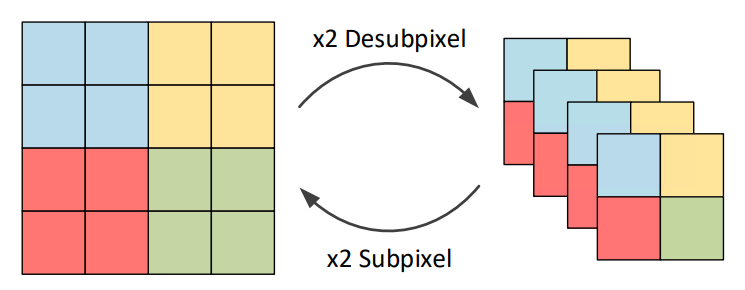}	
	\caption{Subpixel upsampling and desubpixel downsampling. Figure from \cite{feqe}}
	\label{fig:subpixel}
\end{figure}

\paragraph{\textbf{De-raining network}}
In image recovery tasks, downsampling is typically applied to the network to speed up the forward propagation of the network. However, Downsampling generally leads to loss of information which is more severe when performed early in the network.

Considering the speed of forward propagation and the provision of sufficient information for the following convolutional layers, we introduce desubpixel \cite{feqe} instead of common pooling structures and strided convolutions for downsampling, and use the corresponding upsampling operation called subpixel \cite{subpixel_conv} to restore the resolution. Desubpixel is a reversible downsampling module and its input can be recovered by the subpixel \cite{subpixel_conv} as shown in Fig. \ref{fig:subpixel}. In desubpixel, the spatial features are systematically rearranged into channels to reduce spatial dimensions without losing information, keeping the feature values intact, hence providing sufficient information for the following convolutional layers.

A concatenation of operation based on subpixel-upsampling and desubpixel-downsampling will lead to the identity transform such that:

\begin{equation}
U(D(X))=X
\label{up_down_func}
\end{equation}
where U and D respectively denote subpixel-upsampling and desubpixel-downsampling function.

\begin{figure}[hb]
	\centering  
	\includegraphics[width=0.45\textwidth]{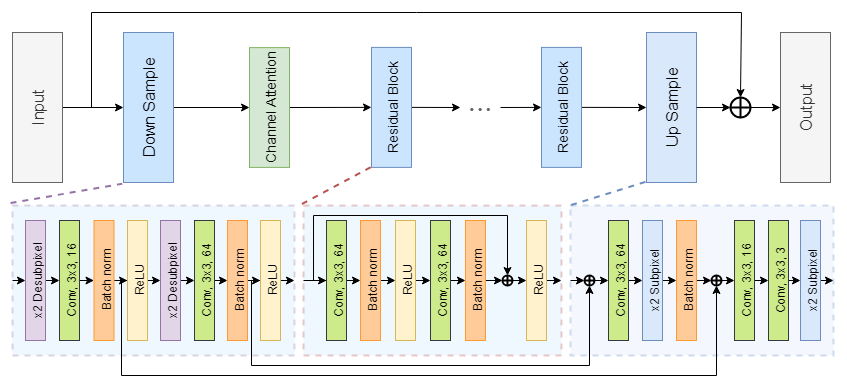}	
	\caption{Detail architecture of the proposed de-raining network.}
	\label{detail_of_deraining_network}
\end{figure}

Based on U-Net\cite{unet}, we propose to use two desubpixel with 2x ratio for the downsampling features extraction and following with several residual blocks, which allows easier to optimize and can gain better performance from considerably increased depth, as shown in Fig. \ref{detail_of_deraining_network}.

On the one hand Channel Attention (CA) Module enables the model to focus on useful features, on the other hand, it reduces the overfitting of features. CA is essentially a feature weight that learns the correlation sympathies and weighting relationships between features, which often provide global guidance.

So we propose to use CA for global feature extraction, and use residual blocks with a 3x3 kernel size for local feature extraction.

\begin{figure}[ht]
	\centering  
	\includegraphics[width=0.45\textwidth]{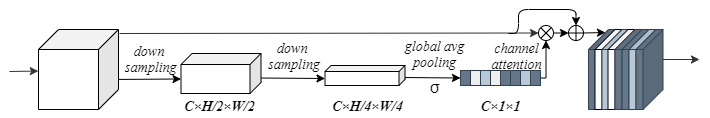}	
	\caption{Detail of Channel Attention (CA) Module architecture. The module computes corresponding attention map through the down-sampling branches. The size of receptive fields which is helpful for the contextual information aggregation at the master branch. "$+$" denotes element-wise addition, $\sigma$ denotes sigmoid activation function, and "$×$" denotes element-wise product. Attention mask is obtained after it is activated by  $\sigma$. Then the short circuit occurs after the channel attention is applied.}
	\label{channel_attention_module}
\end{figure}

In this experiment, we designed the de-rain model structure as shown in above. Note that, the number of residual module layers can be freely adjusted according to the task needs. Typically, in order to improve the prediction accuracy, the network can be designed as deep as possible, because deeper networks tend to provide higher performance than shallow networks. But time consumption also increases as the network deepens.

\paragraph{\textbf{Loss network}}
The rain streak is the higher frequency information relative to surrounding background. Therefore as shown in Fig.7, the edge information of rain streak in fail output of the de-raining network can be spotlighted by loss network and cause larger feature reconstruction loss.

\begin{figure}[ht]
	\centering  
	\includegraphics[width=0.45\textwidth]{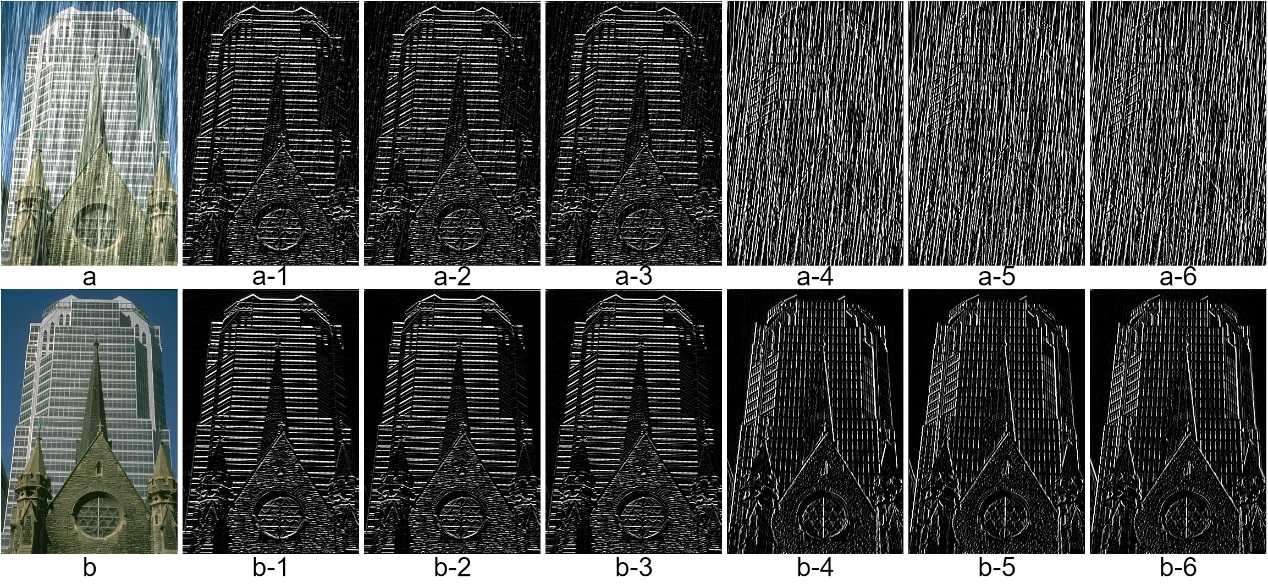}	
	\caption{comparison between spatial domain and gradient domain. (a) and (b) are the RGB images with and without rain respectively. (1) to (6) are their corresponding gradient maps along vertical/horizontal direction (y/x axis).}
	\label{fig:edge_map}
\end{figure}

In addition, as mentioned in \cite{he_guided_filter}, there are mainly two cases in image: high variance areas and flat patches. The flat patches with small variance between pixels are usually smooth and the interior of an object. The image changes a lot within the high variance regions, where the gradients are larger, relatively. Typically, there are rich details and textures in the high variance areas. So in the process of training, edge-aware loss is considered to be penalty for outputs which fail to maintain texture details of targets.

Fig. \ref{fig:edge_map} has shown that certain feature learning architectures also yield useful features for image rain removal task even with untrained, specific weights.

\paragraph{\textbf{Learning Tasks-Constrained DCNN}}

The effectiveness of learning the DCNN by stochastic gradient descent has been proven when a single task is present [32]. However, it is non-trivial to optimize multiple tasks simultaneously using the same method. The reason is that different tasks have different loss functions and learning difficulties, and thus with different convergence rates. In multi-task learning (MTL), how to set appropriate loss weights for each task has always been a problem to be solved. At present, Existing methods [8], [9] involving multi-tasking often manually set a fixed coefficient to each task. They often need time and effort to find a suitable set of coefficients. In order to solve the tedious search of coefficients, we propose a dynamic weighting scheme from the perspective of gradient balance and loss balance.

Unbalanced gradient norms on different tasks will lead to sub-optimal multi-task network training effects. In Algorithm 1, we describe the process of training network models based on the gradient-balanced dynamic weighting (denoted as GB) algorithm. This scheme gets rid of the control of artificial hyper-parameters, and it can achieve the gradient balance in the training process without any additional manual intervention settings. Gradient balance adjusts the loss by calculating a brand new gradient value (by normalizing the gradient size during the backpropagation training process of the network itself). Weights are to resolve imbalances in the gradient norm.

\begin{algorithm}[htb] 
	\algsetup{indent=2em,
		linenosize=\small,
		linenodelimiter=.
	}	
	\caption{\small Gradient-Balanced Task Weighting Algorithm, GB} 
	\textbf{Input:} the losses set of $T$ tasks $\ell$ \\
	\textbf{Output:} The re-weighting loss ${\lambda_{i}*\ell_{i}}$ for each task under constraints. \\ 
	\textbf{Step 1. Initialize:} initialize neural network weights $\mathbf{W}$;\\
	\textbf{Step 2. Evolution:} Find optimal solutions
	
	\begin{algorithmic}[1]
		\FOR{ each epoch $i$}
			\FOR{ each batch of data $B$}
				\STATE Get the loss on each task $t$: $\ell_{(B, t)} \in \mathbb{R}^{T}$;           
				\STATE Get the gradient norm of each task to the shared convolution layer: $g_{(B, t)} \in \mathbb{R}^{T}$;
				\STATE Store the total gradient norm values: $g_{B} \in \mathbb{R}^{T}$;
				\FOR{ each task $t$}
					\STATE Calculate the task weight: $w_{t}=1-g_{(B, t)} / g_{B}$;
					\STATE Update the weighted loss of the current task: $\ell_{(B, t)}=\ell_{(B, t)} \times w_{t}$;
				\ENDFOR		
			\ENDFOR
		\ENDFOR
	\end{algorithmic}
	\textbf{Step 4. Get New total loss:} Store the sum of each re-weighted task loss value: $\ell_{B} \in \mathbb{R}^{T}$ according to problem (\ref{fomula:total_loss}); \\
	\textbf{Step 4. Back propagation:} Update $\mathbf{W}$ with respect to $\ell_{B}$.
	
	\label{alg_gb} 
\end{algorithm}

The coefficients for each task are in fact to balance the impact of the loss of each task on the training process, and to avoid loss values that are too large by some order of magnitude to mask the role of other tasks in promoting the learning process. Loss balance (denoted as LB) is used to adjust the magnitude gap between task losses by calculating new weighted loss values.

\begin{algorithm}[htb] 
	\algsetup{indent=2em,
		linenosize=\small,
		linenodelimiter=.
	}	
	\caption{\small Loss-Balanced Task Weighting Algorithm, LB} 
	\textbf{Input:} the losses set of $T$ tasks $\ell$ \\
	\textbf{Output:} The re-weighting loss ${\lambda_{i}*\ell_{i}}$ for each task under constraints. \\ 
	\textbf{Step 1. Initialize:} initialize neural network weights $\mathbf{W}$;\\
	\textbf{Step 2. Evolution:} Find optimal solutions
	
	\begin{algorithmic}[1]
		\FOR{ each epoch $i$}
		\FOR{ each batch of data $B$}
		\STATE Get the loss on each task $t$: $\ell_{(B, t)} \in \mathbb{R}^{T}$;           
		\STATE Get the current batch total loss as: $\ell_{B} \in \mathbb{R}^{T}$;
		\FOR{ each task $t$}
		\STATE Calculate current task weight: $w_{t}=1-\ell_{(B, t)} / \ell_{B}$;
		\STATE Update the weighted loss of the current task: $\ell_{(B, t)}=\ell_{(B, t)} \times w_{t}$;
		\ENDFOR		
		\ENDFOR
		\ENDFOR
	\end{algorithmic}
	\textbf{Step 4. Get New total loss:} Store the sum of each re-weighted task loss value: $\ell_{B} \in \mathbb{R}^{T}$ according to problem (\ref{fomula:total_loss}); \\
	\textbf{Step 4. Back propagation:} Update $\mathbf{W}$ with respect to $\ell_{B}$.
	
	\label{alg_lb} 
\end{algorithm}

From the description of Algorithm 1, we can see that GB algorithm has the following characteristics: 1. Dynamic task weight 2. Cross-task provides information to balance the training process. 3. The gradient ratio determines the task weight. The loss-balance based dynamic weighting (denoted as LB) algorithm is similar to the GB algorithm. The only difference is the calculation of task weights. We describe the LB algorithm in Algorithm 2.

It should be noted that in theory, a dynamic weighting scheme based on gradient balance should be the best choice, because the loss value is reflected in the model in a gradient manner, and the learning process of the model is implemented according to the gradient descent (rise) method, but this still requires further observation in the experiment.
No matter what kind of scheme, our core idea is to adaptively adjust the weights of multiple tasks, and use these weights to control the real-time gradient of each task, and then back-propagate and update, so as to result in the better balance of multiple tasks in multi-task learning.

\section{Experiments}
Next we provide the qualitative and quantitative evaluations of proposed method are given. We use two evaluation criteria in terms of the difference between each pair of rain-free image and de-rain result. Apart from the widely used structural similarity (SSIM) \cite{ssim} index, we also use an additional evaluation index, namely peak signal-to-noise ratio (PSNR) \cite{psnr}. A higher SSIM indicates the result is closer to the ground truth in terms of image structure properties (SSIM equals 1 for the ground truth). And the higher PSNR is, the better reconstruction is. For quantitative evaluation, we use distribution (mean) PSNR, SSIM to compare the efforts of each component in our methods, and compare our approaches with the techniques proposed by other researchers. For qualitative evaluation, the de-raining results are shown directly. Both two evaluations are on Rain100H/Rain100L test set.

\subsection{Dataset}

Since the pairs of rain and rain-free images of natural scenes are not massively available, here we utilize the synthetic dataset denoted RainH/RainL provided by Yang et al. \cite{jorder}. 

\textbf{Synthetic Data.}
RainL is selected from BSD200 \cite{bsd_dataset} with only one type of rain streaks, which consists of 200 rainy/clean image pairs for training (denoted RainTrainL) and 100 image pairs for testing (denoted Rain100L). Compared with RainL, RainH with five types of streak directions is more challenging, which contains 1800 rainy/clean image pairs for training (denoted RainTrainH) and 100 image pairs for testing (denoted Rain100H).
During data pre-processing, we crop an Image into four corners and the central crop plus the flipped version of these. Consequently, there are 18000 pairs of rain and rain-free patches in our RainTrainH training dataset. 2000 pairs in RainTrainL dataset.

\textbf{Real-World Data.}
Majority of real-world rainy images in our evaluation are from selected SOTA de-raining works. Others are downloaded from google. Anyway they provide diverse contents, including landscapes, cities, faces and so on.

\subsection{Implementation Details}
Comprehensive experiments are performed on synthetic and real rainy images. The optimization method used in our model is the Adam optimizer, with initial learning rate of 1e-3, dividing it by 10 at the 40th epoch, and terminate training at the 100th epoch. The implementation of the proposed de-raining model is conducted on Python3.6, TensorFlow1.8, GeForce GTX TITAN X with 12GB RAM.

\begin{table*}[!htp]
	\centering
	\caption{Ablation study. Average SSIM and PSNR values on synthetic benchmark datasets. Where Fixed indicates that the task weight is manually specified and do not change during training. Otherwise they are adjusted either by LB or GB, which respectively refers to the loss balance weighting and gradient balance weighting algorithm. These results validate the effectiveness of the combination of loss functions and with specified weighting strategy.}
	

	\begin{tabular}{ccccccc|cc|cc}
		\hline
		\multicolumn{7}{c|}{Combination of Components} & \multicolumn{2}{c|}{Rain100L} & \multicolumn{2}{c}{R100H} \\ \hline
		$L_p$ & $L_e$ & $L_t$ & Fixed & GB & LB & CA & SSIM & PSNR & SSIM & PSNR \\ \hline
		\textbf{$\surd$} &  &  &  &  &  &  & 0.9506 & 33.5484 & 0.8146 & 27.5016 \\
		\textbf{$\surd$} & \textbf{$\surd$} &  & \textbf{$\surd$} &  &  &  & 0.9508 & 33.5116 & 0.8200 & 27.6132 \\
		\textbf{$\surd$} & \textbf{$\surd$} &  &  & \textbf{$\surd$} &  &  & 0.9527 & 33.7361 & 0.8457 & 27.8344 \\
		\textbf{$\surd$} & \textbf{$\surd$} &  &  &  & \textbf{$\surd$} &  & 0.9524 & 33.6634 & 0.8393 & 27.9586 \\
		\textbf{$\surd$} & \textbf{$\surd$} & \textbf{$\surd$} &  & \textbf{$\surd$} &  &  & 0.9526 & 33.6836 & 0.8472 & 27.7668 \\
		\textbf{$\surd$} & \textbf{$\surd$} & \textbf{$\surd$} &  &  & \textbf{$\surd$} &  & 0.9512 & 33.5885 & 0.8465 & 27.7977 \\
		\textbf{$\surd$} & \textbf{$\surd$} & \textbf{$\surd$} &  & \textbf{$\surd$} &  & \textbf{$\surd$} & \textbf{0.9554} & \textbf{34.1852} & 0.8524 & 28.1457 \\
		\textbf{$\surd$} & \textbf{$\surd$} & \textbf{$\surd$} &  &  & \textbf{$\surd$} & \textbf{$\surd$} & 0.9548 & 34.0846 & \textbf{0.8534} & \textbf{28.1572} \\ \hline
	\end{tabular}
	\label{tab:ablation_study}
\end{table*}

\subsection{Ablation Experiments}

To see how the model components we present affects performance, we carefully design ablation experiments on the same datasets. Since loss function, weighting algorithm and network are certain tricks we used to get the proposed de-rain system to work, experiments in this study mainly focus on them to have a systematic examination.

We conduct a systematic ablation study on every components while regarding the model trained only by reconstruction error $L_p$ (namely MSE) as baseline. Table \ref{tab:ablation_study} compares the difference of individual component and that of their assembly when running against the same testing data. We note that the proposed method considerably outperforms baseline method in terms of both PSNR and SSIM. Compared with the baseline trained only by $L_p$, other combination strategies have achieved better performance as shown in Table \ref{tab:ablation_study}. To have a close look at the winning combination strategies, we start from comparing the ${L_e} + Fixed$, the ${L_e} + GB$ and the  ${L_e} + LB$ combination. The test results indicate that the model trained by the GB is significantly better than that by the fixed weighting. The fact that the ${L_e} + Fixed$ is defeated reveals that adaptive weighting GB (or LB) is preferable for both Rain100L and Rain100H. When further extending with texture matching loss, the winner is the ${L_e}{L_t} + GB$  combination. In particular, the SSIM indicator can achieve more that 2\% increase when using the proposed GB algorithm. It can be concluded that the proposed adaptive weighting algorithm has definite advantages.

Further we combine the CA component and analyze its role in our de-raining network. We can see that the CA extension leads to more improvement over the network without CA. This reveals that the model?s predictive capability can be further enhanced by carefully-designed deep network, as will be testified again in Tabel \ref{tab:compare_with_other_sota_methods}. Comparing the ${L_e}{L_t} + GB + CA$  and the ${L_e}{L_t} + LB + CA$  strategy, we find that the GB results in the maximal SSIM and the maximal PSNR for the Rain100L dataset while LB outperforms on the Rain100H dataset.

The results of the present ablation study signify that the combination of four components results in a dramatically improved model while individual component also has a certain positive effect.

\begin{figure*}[htp]
	\centering
	\begin{minipage}{8cm}
		\centering
		\includegraphics[width=1\textwidth]{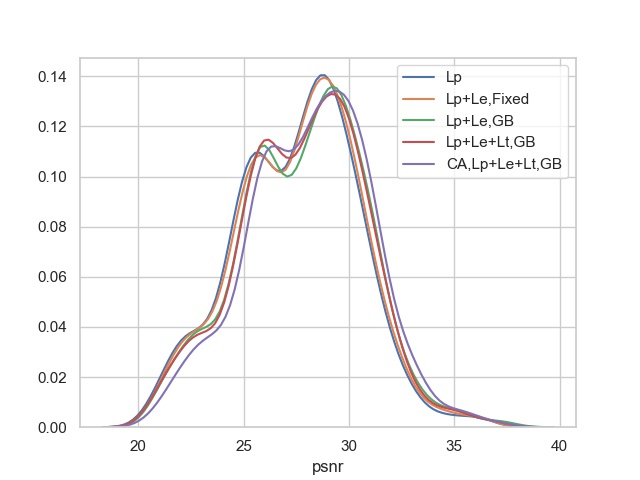}
	\end{minipage}
	\begin{minipage}{8cm}
		\centering
		\includegraphics[width=1\textwidth]{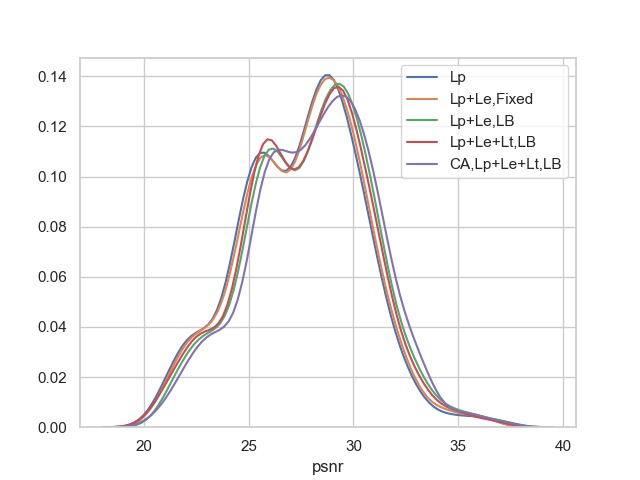}
	\end{minipage}
	\caption{Histogram of per-image PSNR for each combination against Rain100H.}
	\label{fig:psnr_distplot}
\end{figure*}

\begin{figure*}[htp]
	\centering
	\begin{minipage}{8cm}
		\centering
		\includegraphics[width=1\textwidth]{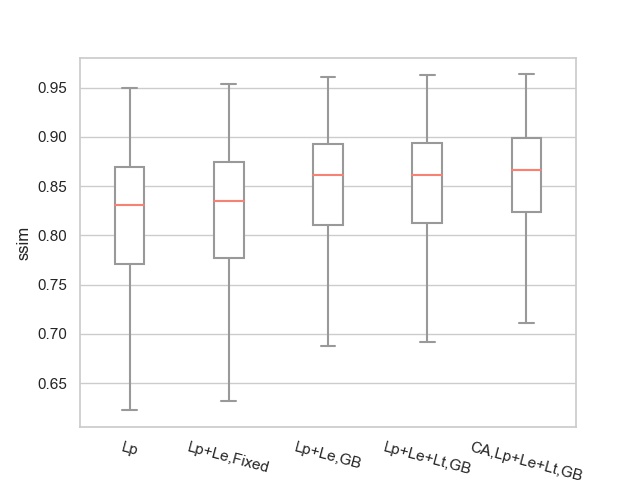}
	\end{minipage}
	\begin{minipage}{8cm}
		\centering
		\includegraphics[width=1\textwidth]{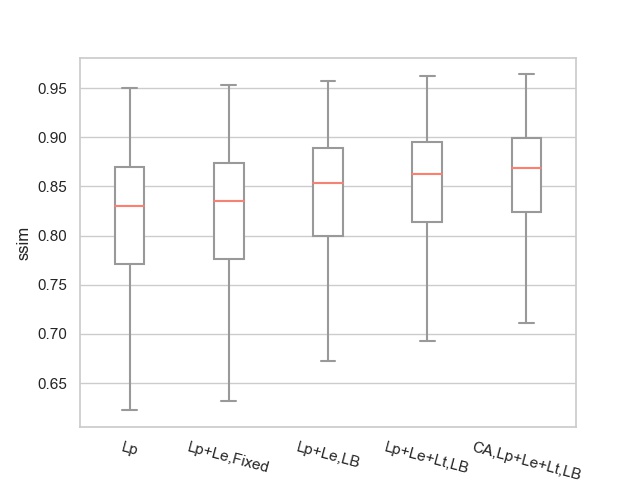}
	\end{minipage}
	\caption{Boxplot of per-image SSIM for each combination against Rain100H.}
	\label{fig:ssim_boxplot}
\end{figure*}

To see how well the results hold up, it is appropriate to examine the distributions of performance in case of random errors. To have an intuitive understanding of how well we control for random errors, we also plot the histogram of per-image PSNR and the boxplot of per-image SSIM for individual component and their combination running against the same dataset Rain100H. For PSNR, Fig.\ref{fig:psnr_distplot} starts from the same baseline $L_p$ and demonstrates two cases with different extensions: one case is the extension through more loss functions without/with GB or CA, the other extends into more loss functions without/with LB or CA. In the first case, the extension from ${L_p},{L_e},{L_p}{L_e} + Fixed,{L_p}{L_e} + GB,{L_p}{L_e}{L_t} + GB$  to ${L_p}{L_e}{L_t} + GB + CA$, leads to the right shift of the distribution curve, especially the distribution of ${L_p}{L_e}{L_t} + GB + CA$ ensures a steady increase compared to others. The same conclusion also holds for the second case (with LB and CA). So we can conclude that the histograms of all four extensions have similar shape and evolve in a positive direction with continued increasing performance. When it comes to the SSIM, Fig.\ref{fig:ssim_boxplot} draws the boxplots of two extensions which resembles Fig.\ref{fig:psnr_distplot}. As revealed by Fig.\ref{fig:ssim_boxplot}, we can conclude that the median of all four extensions for each case increases in a positive direction, and the proposed method has made a significant improvement in the similarity of image structure. Putting them together, the distribution of SSIM/PSNR and the mean SSIM/PSNR shows that the combination of loss functions and adaptive weighting and feed-forward network can bring an overall increase of performance and they have a positive role on de-raining network that figures out the clean images near to ground truth.

In addition to quantitative evaluation, we also presents ablation study results directly that demonstrate the effect of various components in the loss function directly. Comparing the 2nd, 3rd, 4th and 5th images in the Fig. 10, we can observe more and more clear details in the result by constantly strengthening the loss constraint and improving the de-raining network structure.

\begin{figure*}[htp]	
	\centering
	\includegraphics[width=1\textwidth]{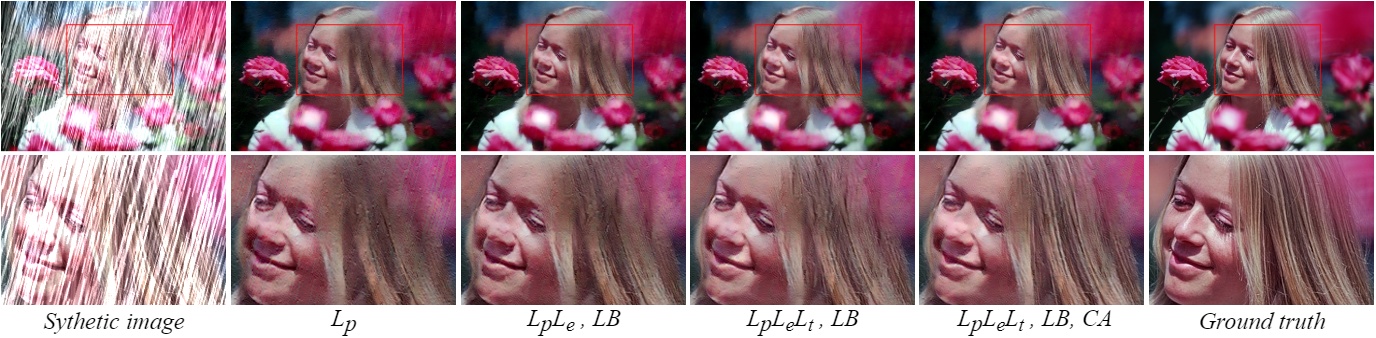}		
	\caption{Ablation study that demonstrates the effectiveness of each component ($L_p,L_e,L_t,CA$), image from RainTrainH. Final result contains more details, distinct contrast, and more natural color.}
\end{figure*}

To quantitatively assess these methods more fairly, we also trained the selected SOTA methods on the same training dataset, and the results are shown in Table \ref{tab:compare_with_other_sota_methods}. For convenience, we record ${L_p}{L_e}{L_t} + GB + CA$ and ${L_p}{L_e}{L_t} + LB + CA$ as MENET-GB and MENET-LB respectively. We can see that our method produces the competitive performance. We used GB/LB tags to distinguish which weighting algorithm our method used for training, and it is important to note that, in order to illustrate the proposed method has ability to achieve or exceed the performance of the SOTA methods, we have showed additional test results for increase in the depth of the de-raining network (increasing the number of residual blocks to 16, and marked with Deep). 


\begin{table*}[htp]
	\centering
	\caption{Quantitative comparison between our method and other four selected SOTA works on synthesized rainy images. The numbers we present here is average SSIM/PSNR. Except for the DSC that does not require training, the other models are re-trained on the corresponding training dataset.}
	\label{tab:compare_with_other_sota_methods}
	\begin{tabular}{c|cc|cc}
		Datasets & \multicolumn{2}{c|}{Rain100L} & \multicolumn{2}{c}{Rain100H} \\ \hline
		Metrics & SSIM & PSNR & \multicolumn{1}{c|}{SSIM} & PSNR \\ \hline
		MENET-GB & 0.9554 & 34.1852 & 0.8524 & 28.1457 \\
		MENET-LB & 0.9548 & 34.0846 & 0.8534 & 28.1572 \\
		MENET-GB-deep & \textbf{0.9559} & \textbf{34.2239} & {\color[HTML]{FE0000} \textbf{0.8610}} & {\color[HTML]{FE0000} \textbf{28.7399}} \\
		MENET-LB-deep & 0.9556 & 34.1839 & 0.8591 & 28.7167 \\ \hline
		DSC & 0.8275 & 25.4473 & 0.5530 & 20.2846 \\
		DDN & 0.9082 & 30.1555 & 0.7400 & 24.2562 \\
		JORDER-E & {\color[HTML]{FE0000} \textbf{0.9792}} & {\color[HTML]{FE0000} \textbf{37.3676}} & 0.8555 & 27.7527 \\
		RESCAN & 0.9672 & 35.5393 & \textbf{0.8572} & \textbf{28.2138} \\ \hline
	\end{tabular}
\end{table*}

\begin{figure*}[hb]	
	\centering
	\includegraphics[width=1\textwidth]{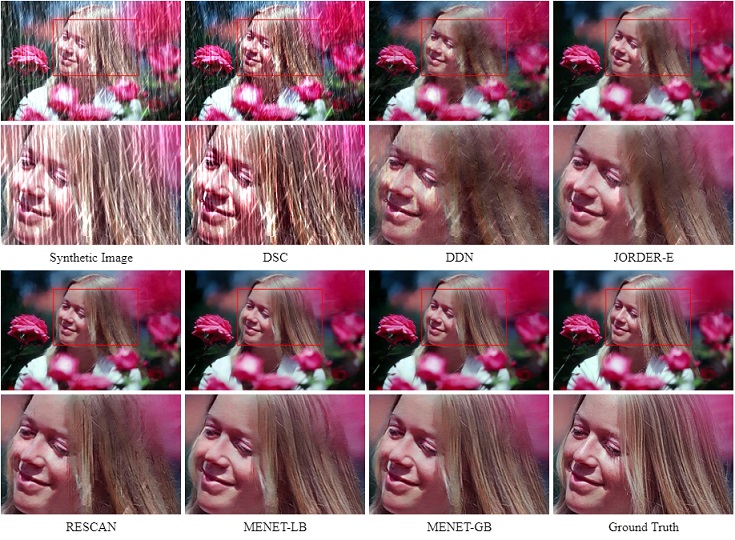}		
	\caption{Visual comparison of MENET with other methods on synthetic data.}
	\label{fig:compare_texture_detail}
\end{figure*}

As observed in Fig. \ref{fig:compare_texture_detail}, there is rain streak obviously remained in the output of DSC. JORDER-E tends to over-smooth the image content. DDN occasionally fails to capture the rain streak. It can be seen that our results are closer to the ground truth as indicated by SSIM and PSNR in Table \ref{tab:compare_with_other_sota_methods}. The indices in Table \ref{tab:compare_with_other_sota_methods} support the competitive advantage of our methods.

\begin{table*}[htbp]
	\centering
	\caption{Comparison of running time (seconds/image). CPU (Inter Pentium CPU G645 @2.9GHz, RAM 6.00GB)/GPU (GeForce GTX TITAN X)}	
	\begin{tabular}{l|lllll}
		\toprule
		Image size & \multicolumn{1}{c}{Ours} & \multicolumn{1}{c}{DSC} & \multicolumn{1}{c}{DDN} & \multicolumn{1}{c}{JORDER-E} & \multicolumn{1}{c}{RESCAN} \\
		\midrule
		224x224 & 0.6578/0.0030 & 70.4289 / - & 1.8226/0.0084 & 27.6392/0.0151 & 4.2296/0.0403 \\
		\bottomrule
	\end{tabular}%
	\label{tab:run_time}%
\end{table*}%

Compare with other non-deep methods, ours proposed approach can process new images very efficiently. Table \ref{tab:run_time} shows the average running time of processing a test image for 224x224 image sizes, each averaged over 1000 testing images. DDN, JORDER-E and RESCAN are implemented using both CPUs and GPUs according to the provided source code, while our method is tested on both CPUs and GPUs. Since method (DSC) is based on dictionary learning, complex optimizations are still required to de-raining test images, which accounts for the slower computation time. RESCAN is not very fast because it contains multiple stages of cycle. DDN gets faster speed due to small convolution channels. As observed from results, ours method is remarkably faster than any other methods listed in Table \ref{tab:run_time}, this is due to our downsampling operation reducing the spatial dimension of the feature.

\begin{table*}[htp]
	\centering
	\caption{Comparison of VGG19 loss network and Edge-aware loss network. We use the feature maps obtained by the forth convolution before the third max-pooling layer of VGG19.}
	\label{tab:compare_with_vgg}
	\begin{tabular}{cc|cc|cc}
		\hline
		&  & \multicolumn{2}{c|}{Rain100L} & \multicolumn{2}{c}{Rain100H} \\ \hline
		\multicolumn{1}{c|}{Components} & Coefficients & SSIM & PSNR & SSIM & PSNR \\ \hline
		\multicolumn{1}{c|}{MSE (Baseline)} & / & 0.9506 & 33.5484 & 0.8146 & 27.5016 \\
		\multicolumn{1}{c|}{Fixed, MSE, VGG} & 1e-3 & 0.9550 & 33.8075 & 0.8210 & 27.6265 \\
		\multicolumn{1}{c|}{Fixed, MSE, Edge} & 1e-2 & 0.9508 & 33.5116 & 0.8200 & 27.6132 \\ \hline
	\end{tabular}
\end{table*}

It is worth mentioning that VGG as loss network requires more physical resources (more than twice as much as the edge-aware loss network), in addition, its FLOPs (floating point operations) is much larger than the edge-aware loss network, resulting in a much higher training time than the Edge-aware loss network. In our works, application of edge-aware loss network considered to be time efficient, effective and simple. So we simply report the behavior of untrained edge-aware loss networks and off-the-shelf VGG19 loss network, the results are presented in Table \ref{tab:compare_with_vgg}. Although we can see from Table \ref{tab:compare_with_vgg} that VGG19 is slightly more advantageous than edge as loss network, in our works, application of edge-aware loss network considered to be time efficient, effective and simple.

\textbf{Results.}
As shown in Table \ref{tab:ablation_study}, our model achieves the superior result. From these comparison, we can see the key roles of edge-aware loss and texture matching loss. Owing to taking advantage of the synergy learning and well-designed network, our methods performs competitive performants with others.

\begin{figure*}[htp]
    \centering
     \begin{minipage}{5cm}
        \centering
        \includegraphics[width=1\textwidth]{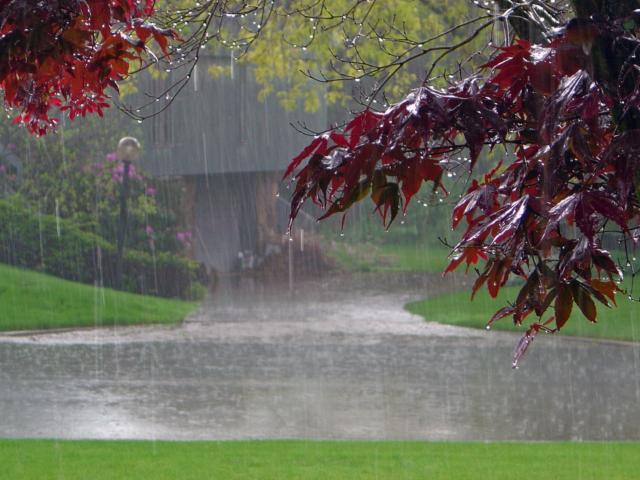}
    \end{minipage}
    \begin{minipage}{5cm}
        \centering
        \includegraphics[width=1\textwidth]{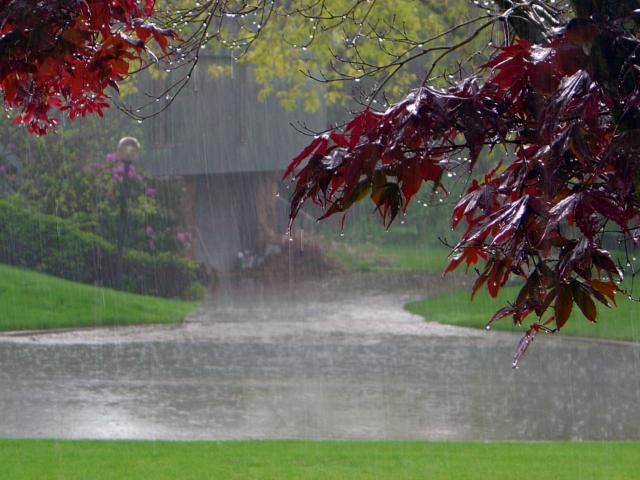}
    \end{minipage}
    \begin{minipage}{5cm}
        \centering
        \includegraphics[width=1\textwidth]{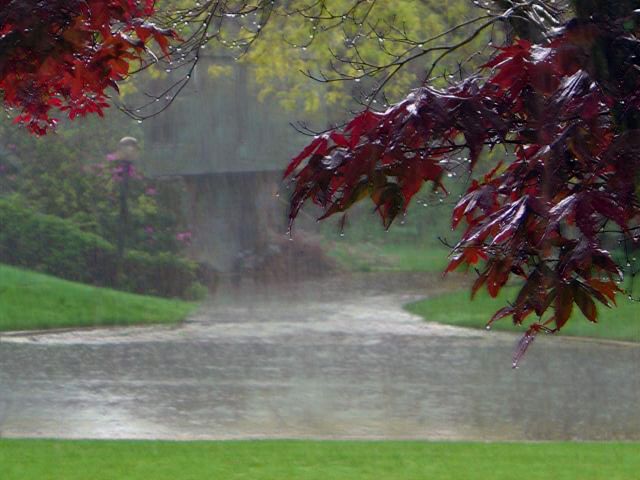}
    \end{minipage}  
    
    \vspace{.15cm}
      
    \begin{minipage}{5cm}
        \centering
        \includegraphics[width=1\textwidth]{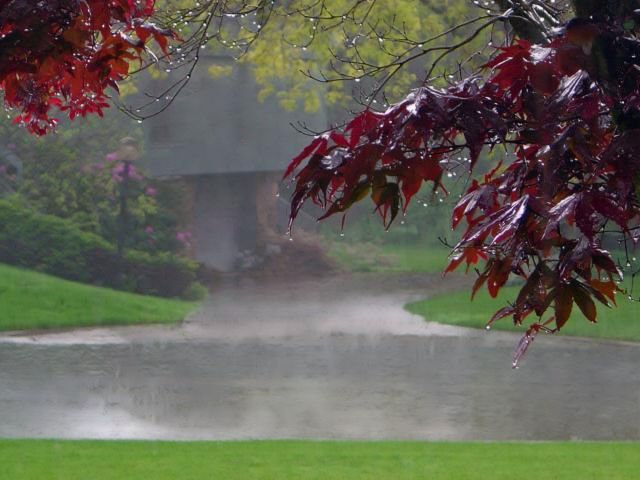}
    \end{minipage}
    \begin{minipage}{5cm}
        \centering
        \includegraphics[width=1\textwidth]{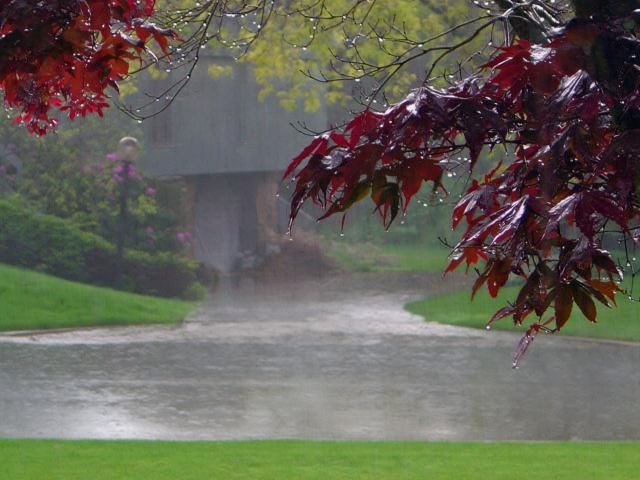}
    \end{minipage}
    \begin{minipage}{5cm}
        \centering
        \includegraphics[width=1\textwidth]{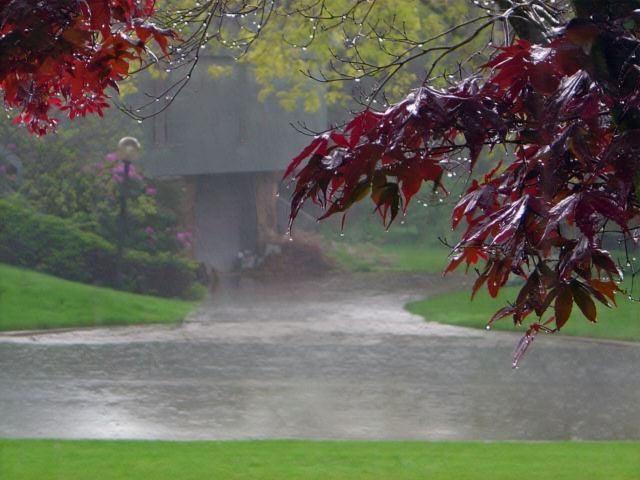}
    \end{minipage}
    
    \vspace{.2cm}
    
    \begin{minipage}{5cm}
    	\centering
    	\includegraphics[width=1\textwidth]{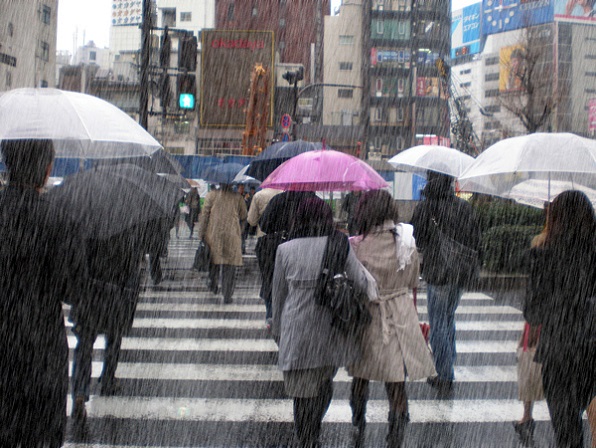}
    \end{minipage}
    \begin{minipage}{5cm}
    	\centering
    	\includegraphics[width=1\textwidth]{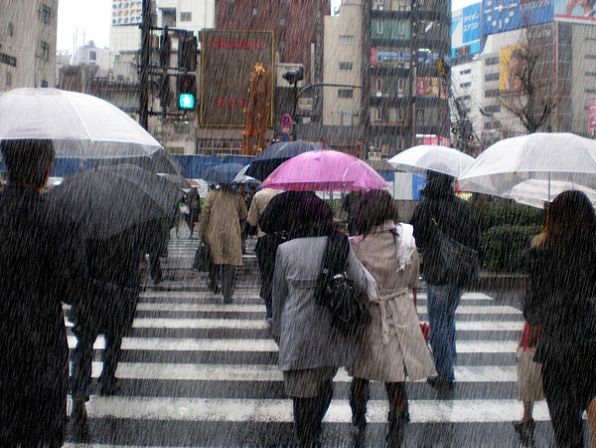}
    \end{minipage}
    \begin{minipage}{5cm}
    	\centering
    	\includegraphics[width=1\textwidth]{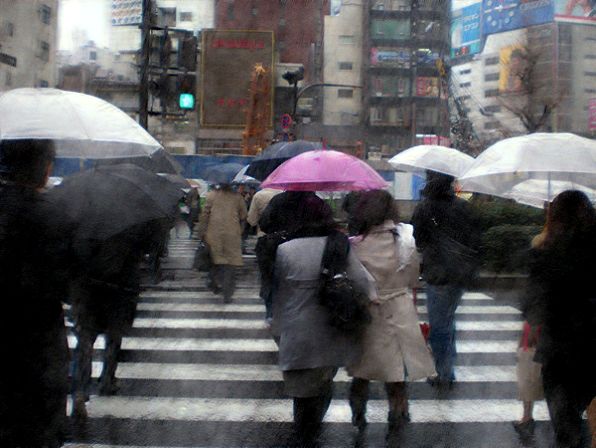}
    \end{minipage}  
    
    \vspace{.15cm}
    
    \begin{minipage}{5cm}
    	\centering
    	\includegraphics[width=1\textwidth]{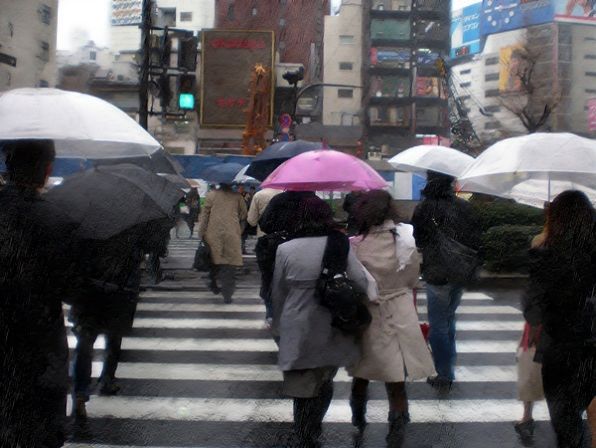}
    \end{minipage}
    \begin{minipage}{5cm}
    	\centering
    	\includegraphics[width=1\textwidth]{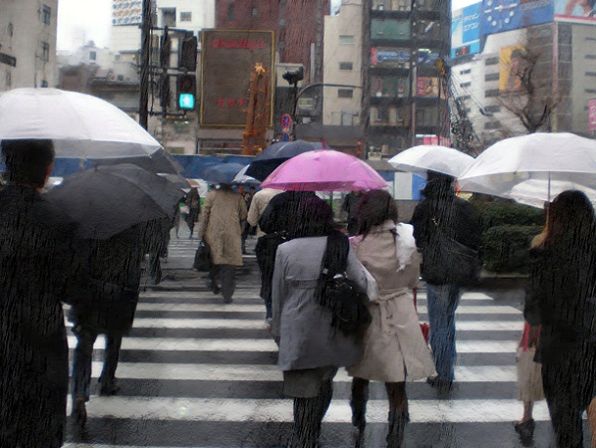}
    \end{minipage}
    \begin{minipage}{5cm}
    	\centering
    	\includegraphics[width=1\textwidth]{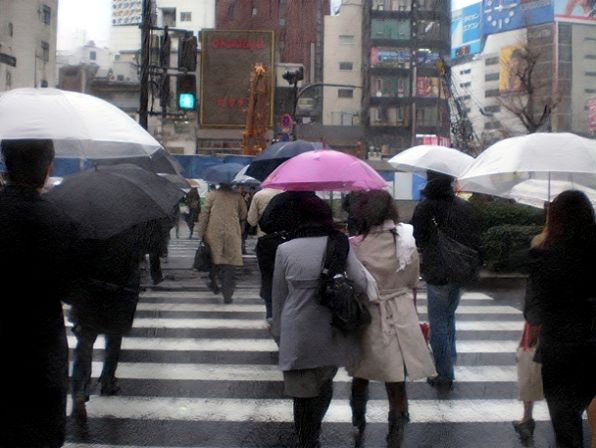}
    \end{minipage}

	\vspace{.2cm}
	
	\begin{minipage}{5cm}
		\centering
		\includegraphics[width=1\textwidth]{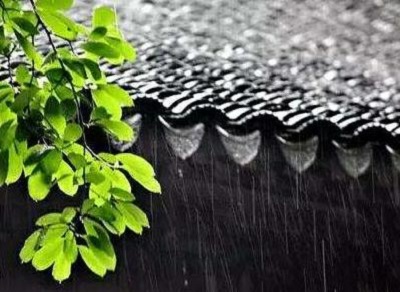}
	\end{minipage}
	\begin{minipage}{5cm}
		\centering
		\includegraphics[width=1\textwidth]{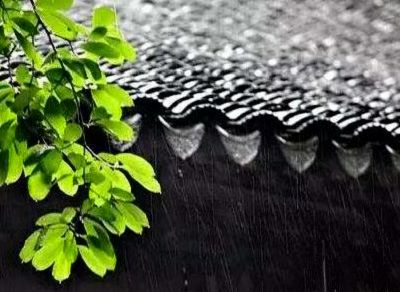}
	\end{minipage}
	\begin{minipage}{5cm}
		\centering
		\includegraphics[width=1\textwidth]{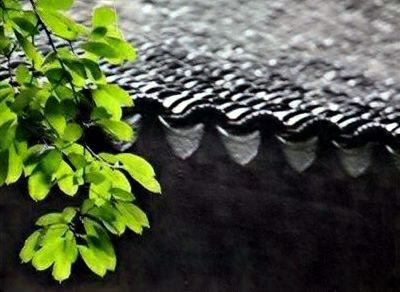}
	\end{minipage}  
	
	\vspace{.15cm}
	
	\begin{minipage}{5cm}
		\centering
		\includegraphics[width=1\textwidth]{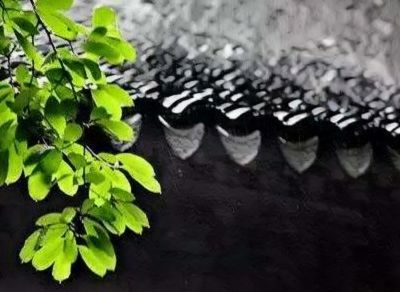}
	\end{minipage}
	\begin{minipage}{5cm}
		\centering
		\includegraphics[width=1\textwidth]{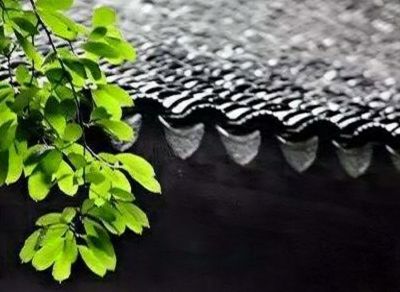}
	\end{minipage}
	\begin{minipage}{5cm}
		\centering
		\includegraphics[width=1\textwidth]{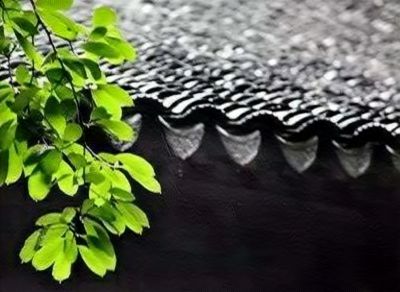}
	\end{minipage}
    
    \caption{Light rain, left to right, top to down are rainy image, DSC, DDN, JORDER-E, RESCAN and MENET-LB.}
    \label{fig:light_derain}
\end{figure*}

\begin{figure*}[htp]
	\centering
	\begin{minipage}{5cm}
		\centering
		\includegraphics[width=1\textwidth]{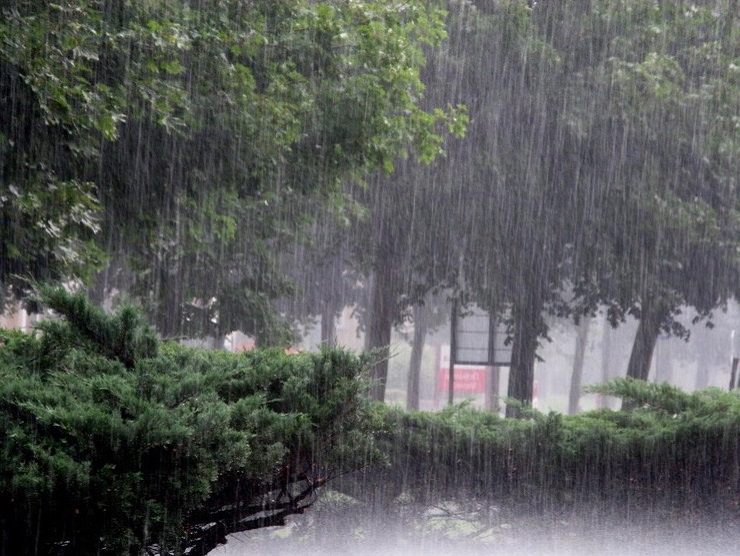}
	\end{minipage}
	\begin{minipage}{5cm}
		\centering
		\includegraphics[width=1\textwidth]{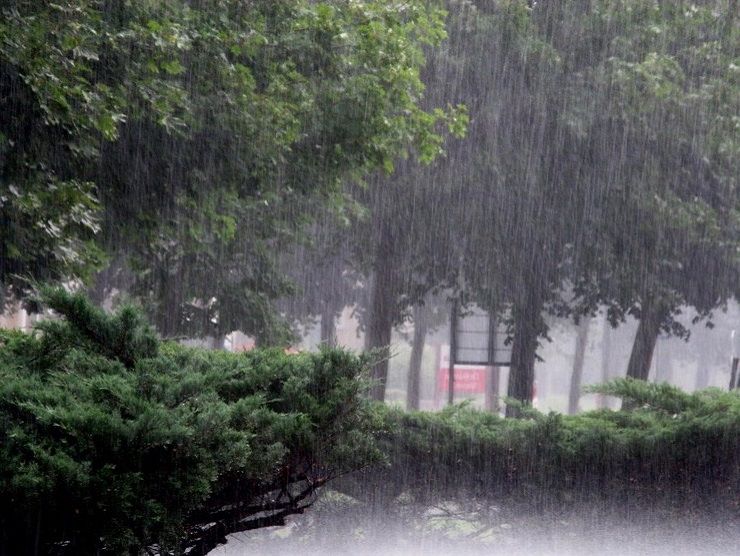}
	\end{minipage}
	\begin{minipage}{5cm}
		\centering
		\includegraphics[width=1\textwidth]{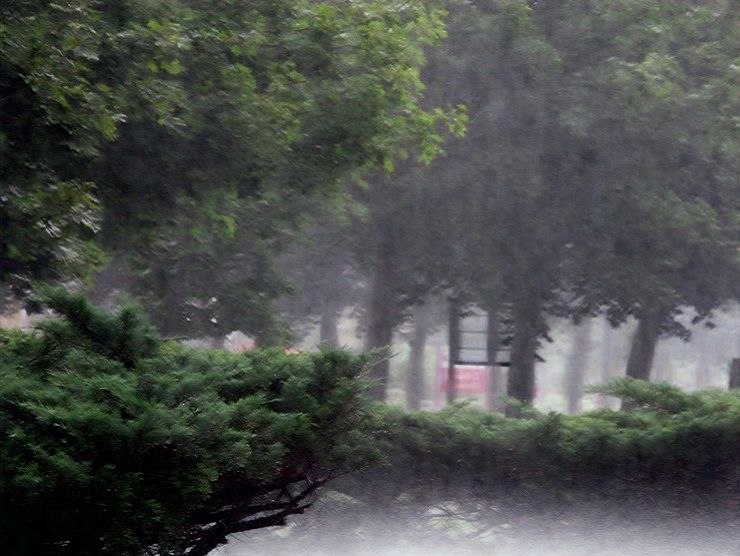}
	\end{minipage}  
	
	\vspace{.15cm}
	
	\begin{minipage}{5cm}
		\centering
		\includegraphics[width=1\textwidth]{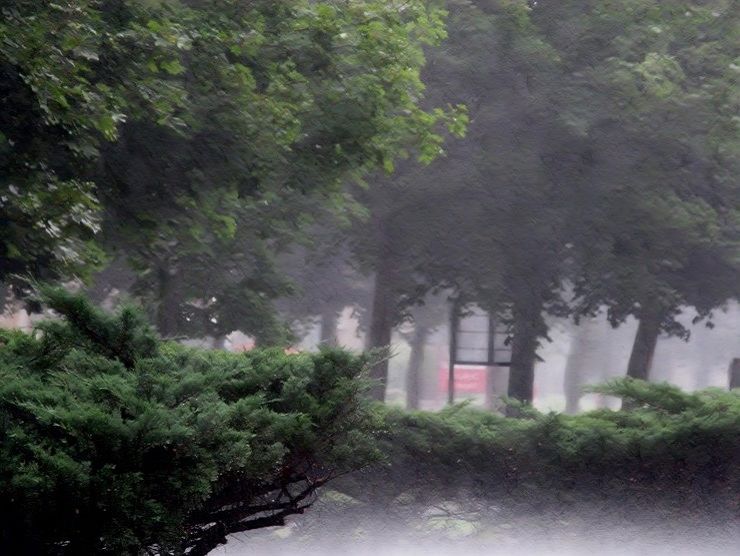}
	\end{minipage}
	\begin{minipage}{5cm}
		\centering
		\includegraphics[width=1\textwidth]{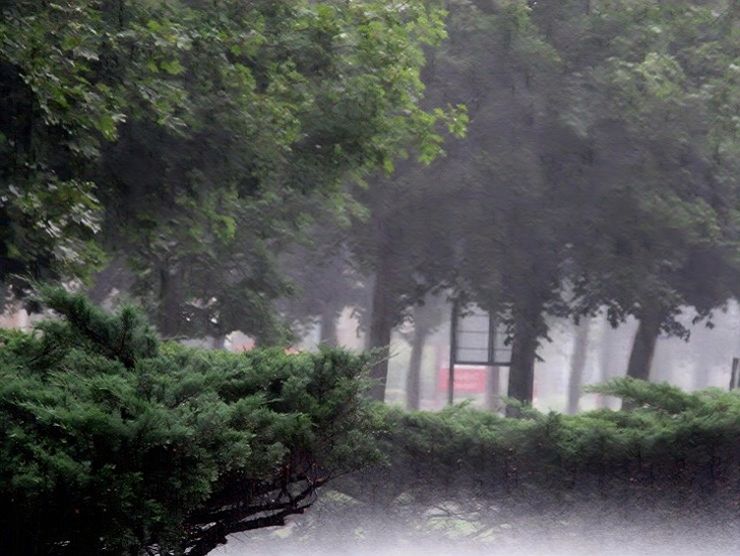}
	\end{minipage}
	\begin{minipage}{5cm}
		\centering
		\includegraphics[width=1\textwidth]{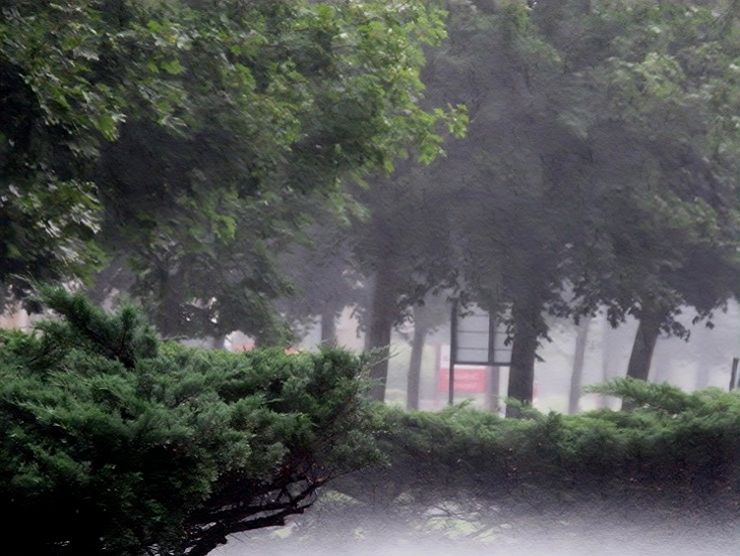}
	\end{minipage}
	
	\vspace{.2cm}
	
	\begin{minipage}{5cm}
		\centering
		\includegraphics[width=1\textwidth]{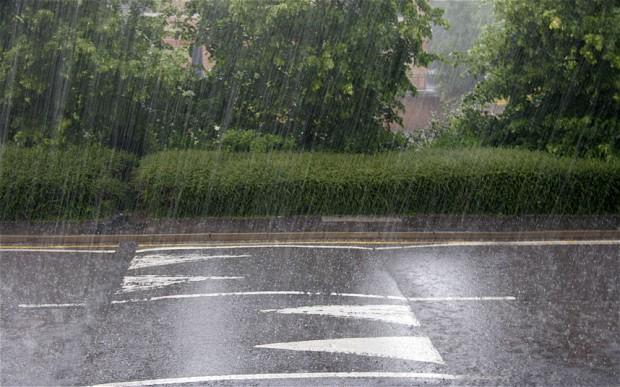}
	\end{minipage}
	\begin{minipage}{5cm}
		\centering
		\includegraphics[width=1\textwidth]{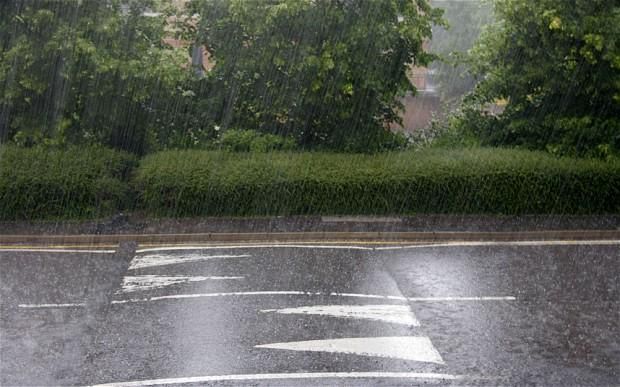}
	\end{minipage}
	\begin{minipage}{5cm}
		\centering
		\includegraphics[width=1\textwidth]{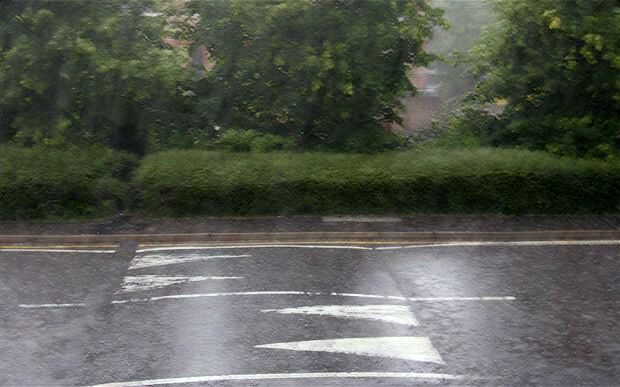}
	\end{minipage}  
	
	\vspace{.15cm}
	
	\begin{minipage}{5cm}
		\centering
		\includegraphics[width=1\textwidth]{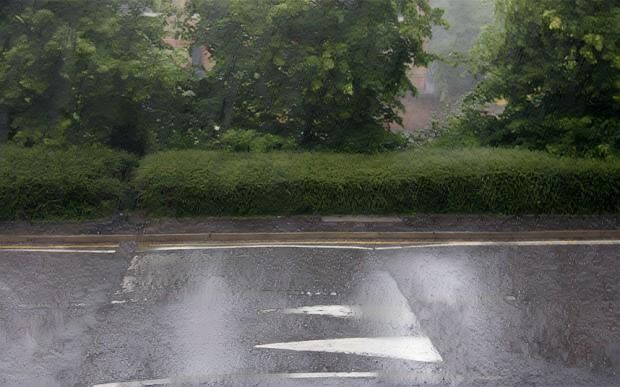}
	\end{minipage}
	\begin{minipage}{5cm}
		\centering
		\includegraphics[width=1\textwidth]{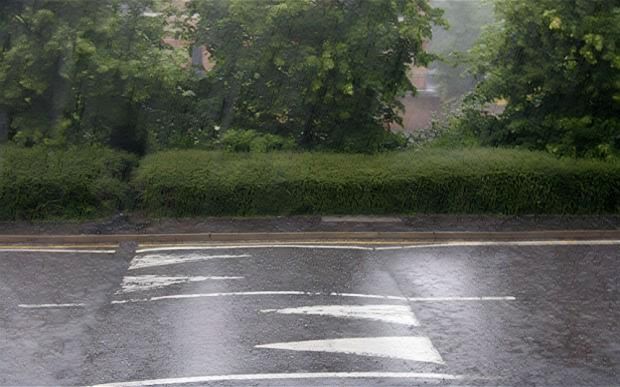}
	\end{minipage}
	\begin{minipage}{5cm}
		\centering
		\includegraphics[width=1\textwidth]{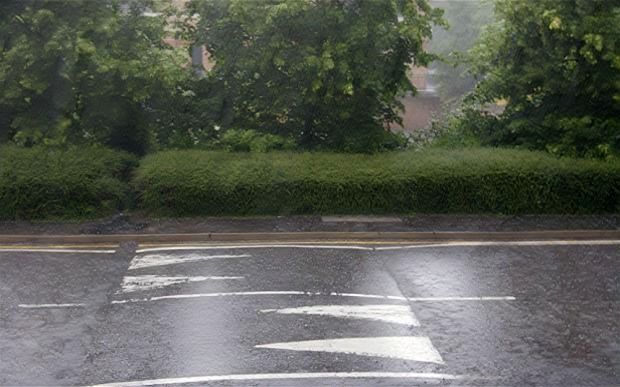}
	\end{minipage}
	
	\vspace{.2cm}
	
	\begin{minipage}{5cm}
		\centering
		\includegraphics[width=1\textwidth]{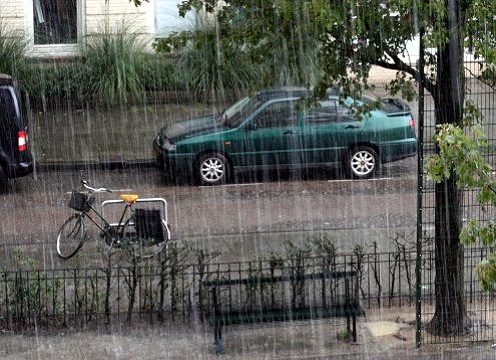}
	\end{minipage}
	\begin{minipage}{5cm}
		\centering
		\includegraphics[width=1\textwidth]{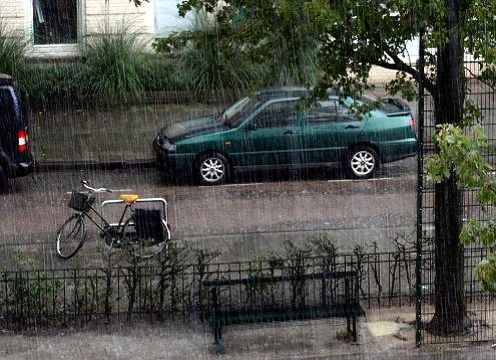}
	\end{minipage}
	\begin{minipage}{5cm}
		\centering
		\includegraphics[width=1\textwidth]{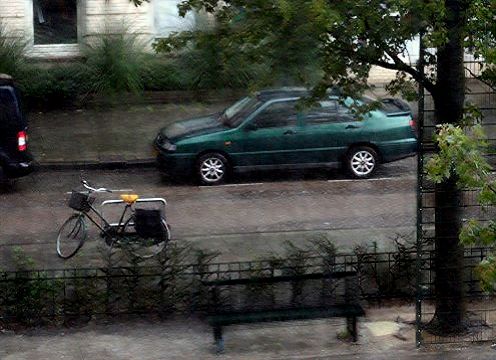}
	\end{minipage}  
	
	\vspace{.15cm}
	
	\begin{minipage}{5cm}
		\centering
		\includegraphics[width=1\textwidth]{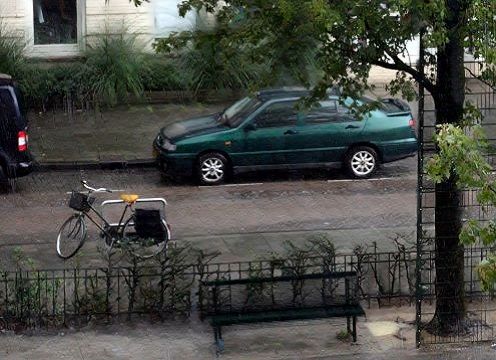}
	\end{minipage}
	\begin{minipage}{5cm}
		\centering
		\includegraphics[width=1\textwidth]{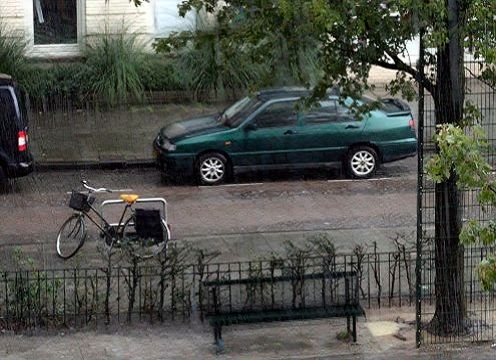}
	\end{minipage}
	\begin{minipage}{5cm}
		\centering
		\includegraphics[width=1\textwidth]{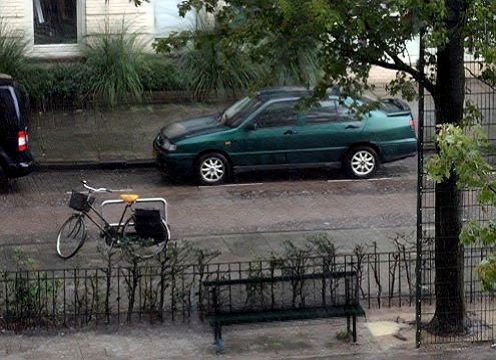}
	\end{minipage}

	\caption{Moderate rain, left to right, top to down are rainy image, DSC, DDN, JORDER-E, RESCAN and MENET-LB.}
	\label{fig:moderate_derain}
\end{figure*}

\begin{figure*}[htp]
	\centering
	\begin{minipage}{5cm}
		\centering
		\includegraphics[width=1\textwidth]{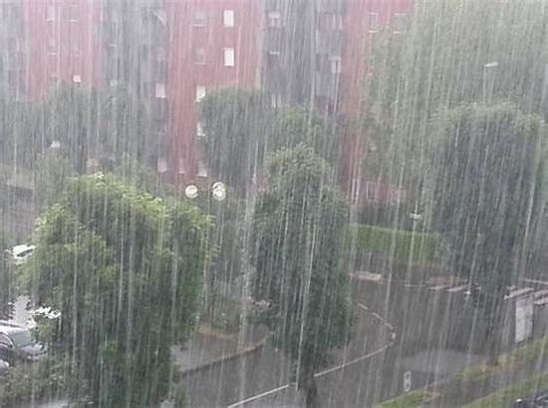}
	\end{minipage}
	\begin{minipage}{5cm}
		\centering
		\includegraphics[width=1\textwidth]{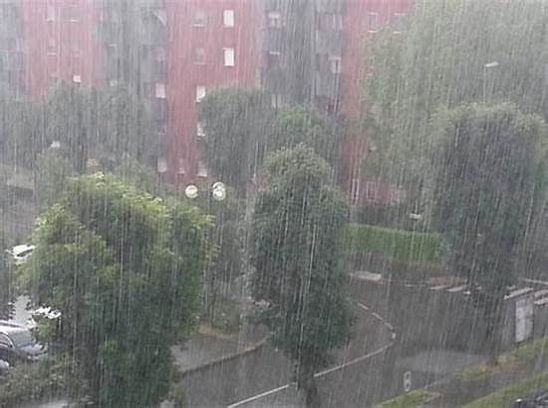}
	\end{minipage}
	\begin{minipage}{5cm}
		\centering
		\includegraphics[width=1\textwidth]{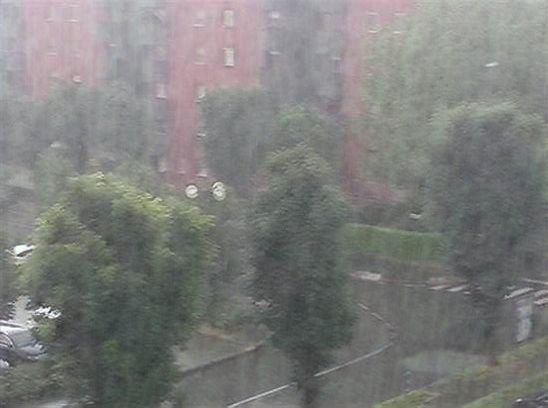}
	\end{minipage}  
	
	\vspace{.15cm}
	
	\begin{minipage}{5cm}
		\centering
		\includegraphics[width=1\textwidth]{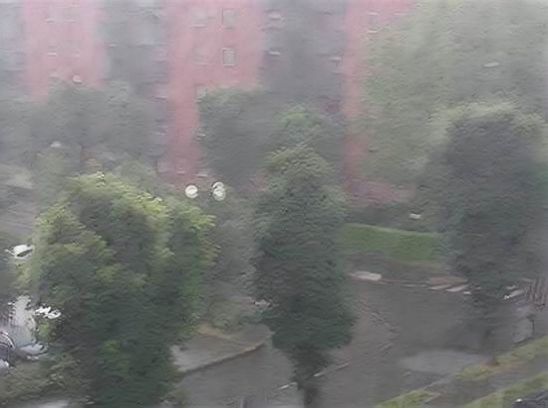}
	\end{minipage}
	\begin{minipage}{5cm}
		\centering
		\includegraphics[width=1\textwidth]{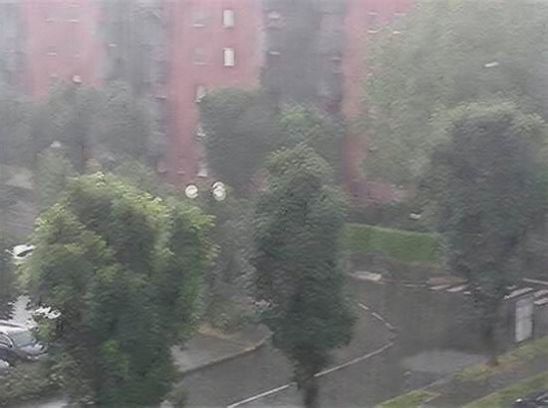}
	\end{minipage}
	\begin{minipage}{5cm}
		\centering
		\includegraphics[width=1\textwidth]{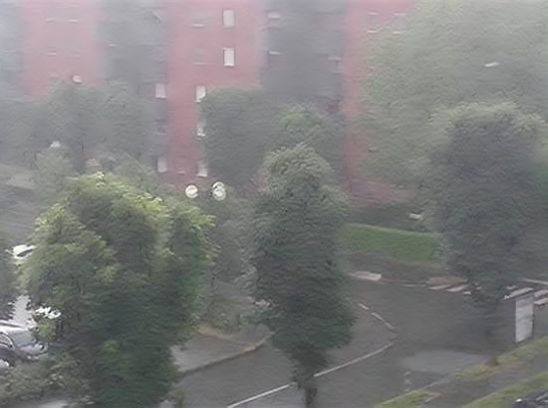}
	\end{minipage}
	
	\vspace{.2cm}
	
	\begin{minipage}{5cm}
		\centering
		\includegraphics[width=1\textwidth]{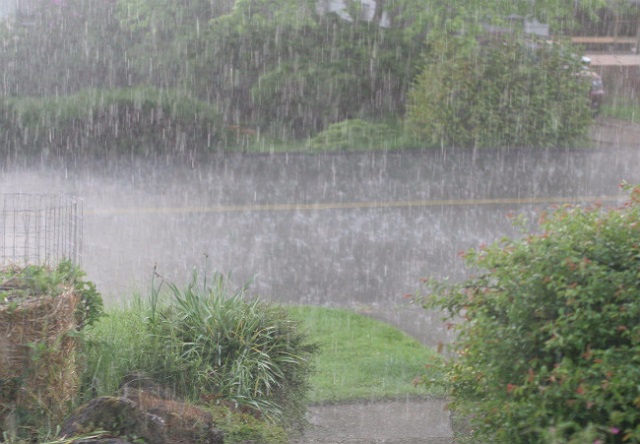}
	\end{minipage}
	\begin{minipage}{5cm}
		\centering
		\includegraphics[width=1\textwidth]{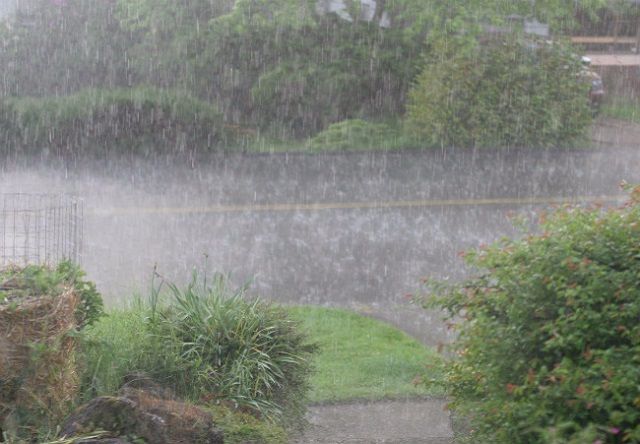}
	\end{minipage}
	\begin{minipage}{5cm}
		\centering
		\includegraphics[width=1\textwidth]{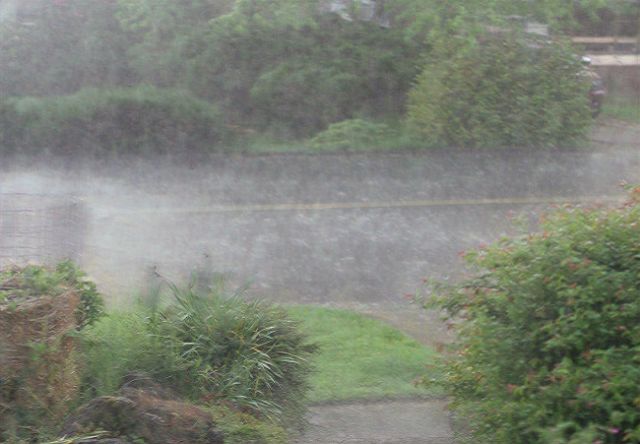}
	\end{minipage}  
	
	\vspace{.15cm}
	
	\begin{minipage}{5cm}
		\centering
		\includegraphics[width=1\textwidth]{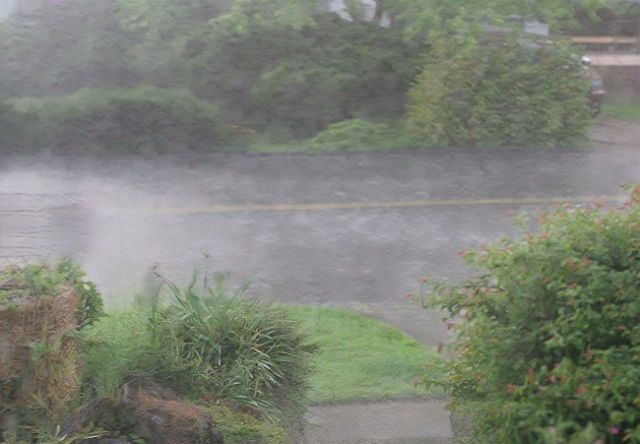}
	\end{minipage}
	\begin{minipage}{5cm}
		\centering
		\includegraphics[width=1\textwidth]{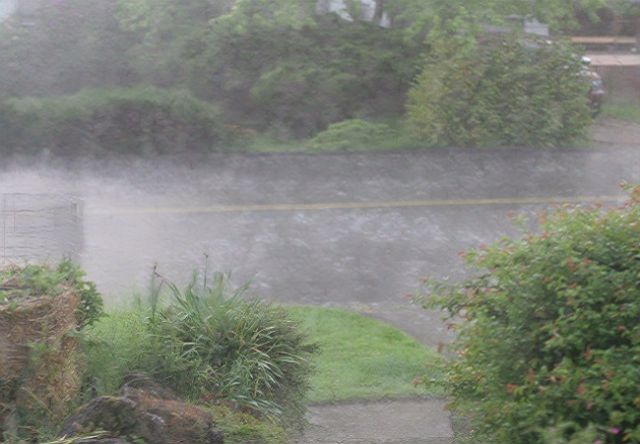}
	\end{minipage}
	\begin{minipage}{5cm}
		\centering
		\includegraphics[width=1\textwidth]{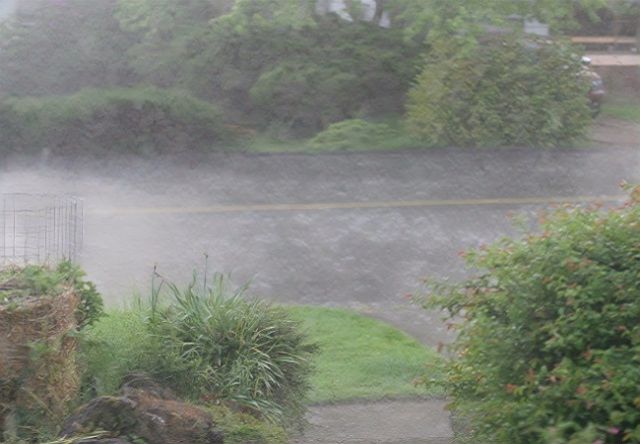}
	\end{minipage}
	
	\vspace{.2cm}
	
	\begin{minipage}{5cm}
		\centering
		\includegraphics[width=1\textwidth]{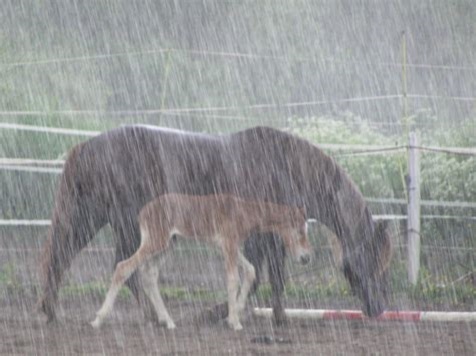}
	\end{minipage}
	\begin{minipage}{5cm}
		\centering
		\includegraphics[width=1\textwidth]{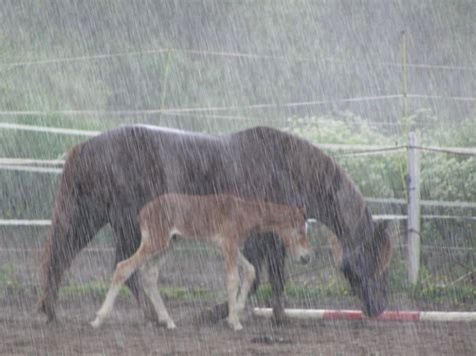}
	\end{minipage}
	\begin{minipage}{5cm}
		\centering
		\includegraphics[width=1\textwidth]{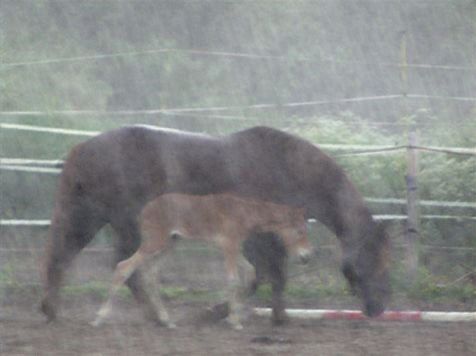}
	\end{minipage}  
	
	\vspace{.15cm}
	
	\begin{minipage}{5cm}
		\centering
		\includegraphics[width=1\textwidth]{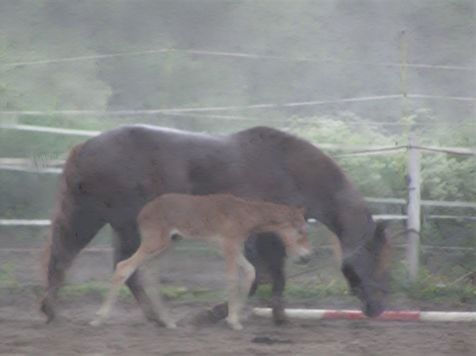}
	\end{minipage}
	\begin{minipage}{5cm}
		\centering
		\includegraphics[width=1\textwidth]{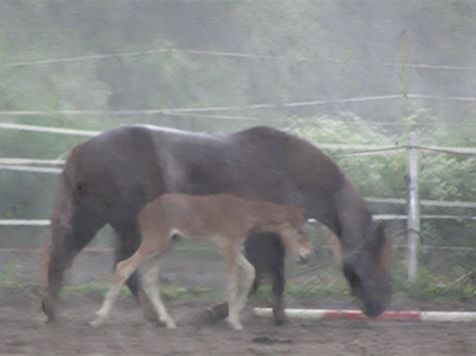}
	\end{minipage}
	\begin{minipage}{5cm}
		\centering
		\includegraphics[width=1\textwidth]{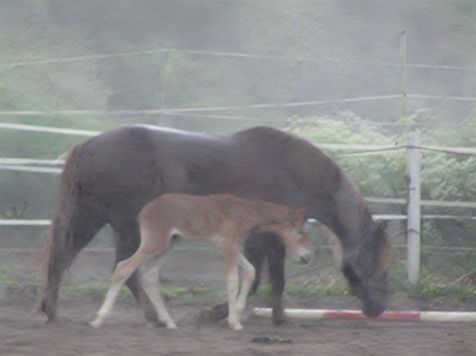}
	\end{minipage}
	
	\caption{Heavy rain, in this case rainy image usually with light fog. left to right, top to down are rainy image, DSC, DDN, JORDER-E, RESCAN and MENET-LB.}
	\label{fig:heavy_derain}
\end{figure*}

\begin{figure*}[htp]
    \centering
     	\begin{minipage}{5cm}
     	\centering
     	\includegraphics[width=1\textwidth]{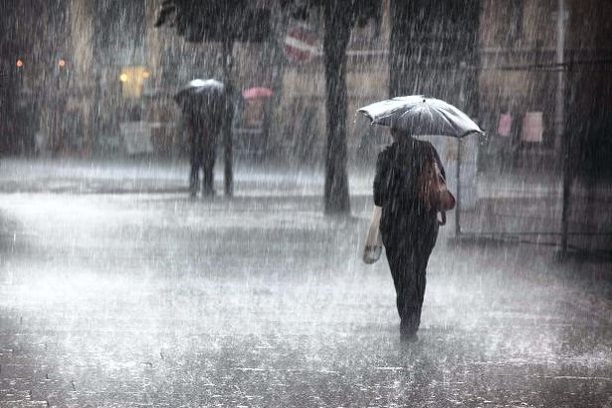}
     \end{minipage}
     \begin{minipage}{5cm}
     	\centering
     	\includegraphics[width=1\textwidth]{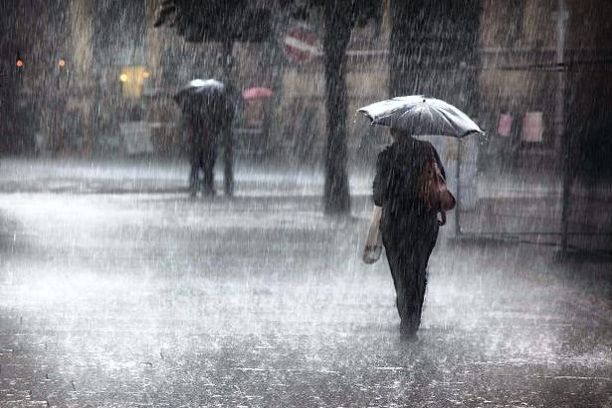}
     \end{minipage}
     \begin{minipage}{5cm}
     	\centering
     	\includegraphics[width=1\textwidth]{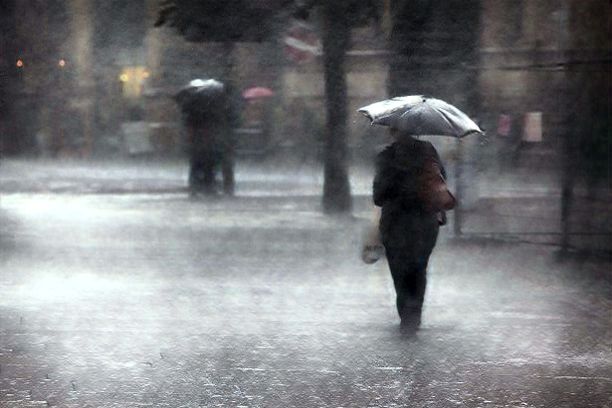}
     \end{minipage}  
     
     \vspace{.15cm}
     
     \begin{minipage}{5cm}
     	\centering
     	\includegraphics[width=1\textwidth]{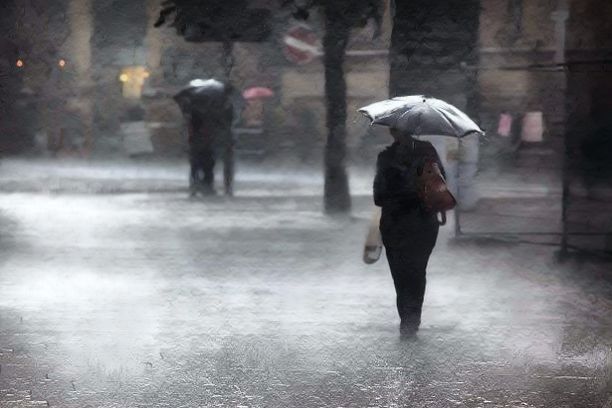}
     \end{minipage}
     \begin{minipage}{5cm}
     	\centering
     	\includegraphics[width=1\textwidth]{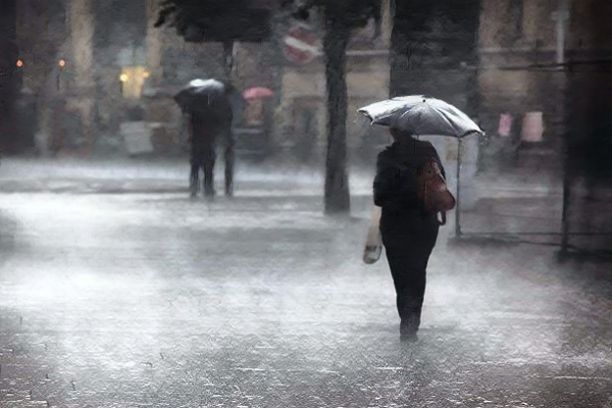}
     \end{minipage}
     \begin{minipage}{5cm}
     	\centering
     	\includegraphics[width=1\textwidth]{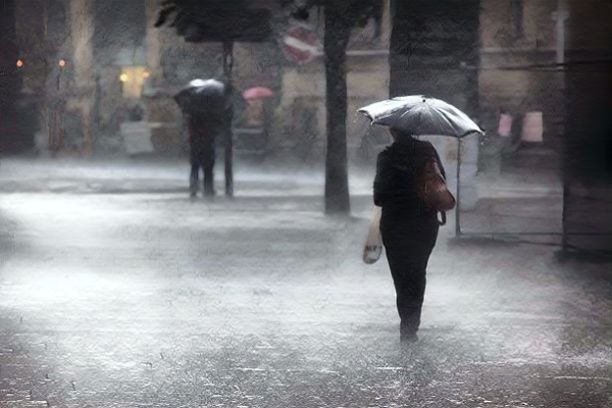}
     \end{minipage}
    
    \caption{ Strong light example, left to right, top to down are rainy image, DSC, DDN, JORDER-E, RESCAN and MENET-LB.}
    \label{fig:strong_light_derain}
\end{figure*}

\begin{figure*}[htp]
    \centering
    \begin{minipage}{5cm}
    	\centering
    	\includegraphics[width=1\textwidth]{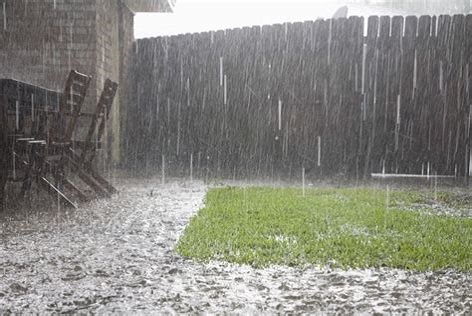}
    \end{minipage}
    \begin{minipage}{5cm}
    	\centering
    	\includegraphics[width=1\textwidth]{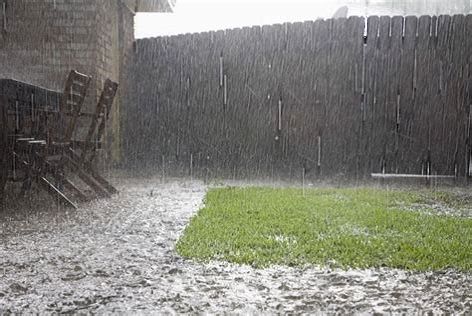}
    \end{minipage}
    \begin{minipage}{5cm}
    	\centering
    	\includegraphics[width=1\textwidth]{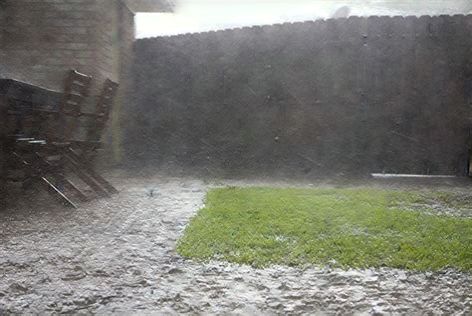}
    \end{minipage}  
    
    \vspace{.15cm}
    
    \begin{minipage}{5cm}
    	\centering
    	\includegraphics[width=1\textwidth]{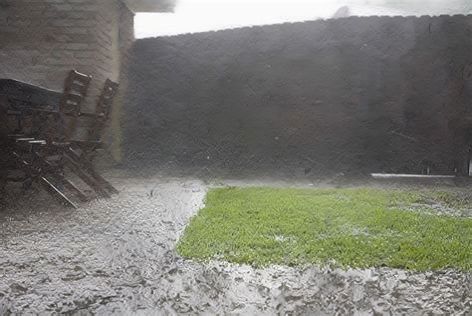}
    \end{minipage}
    \begin{minipage}{5cm}
    	\centering
    	\includegraphics[width=1\textwidth]{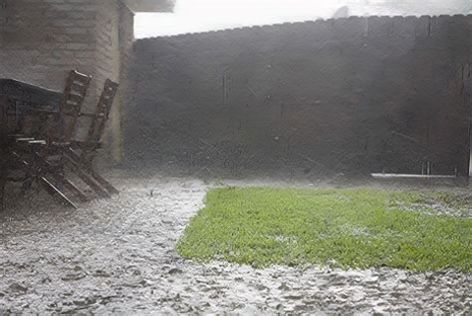}
    \end{minipage}
    \begin{minipage}{5cm}
    	\centering
    	\includegraphics[width=1\textwidth]{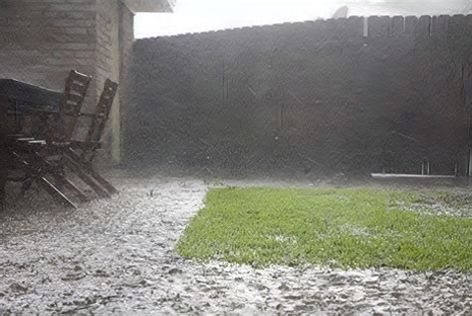}
    \end{minipage}
    
    \caption{Example of failure, left to right, top to down are rainy image, DSC, DDN, JORDER-E, RESCAN and MENET-LB.}
    \label{fig:railure_derain}
\end{figure*}

\subsection{Results on real-world rainy images}

In this section, we will directly demonstrate the effects of several selected methods(DSC\cite{dsc}, DDN\cite{ddn}, JORDER-E\cite{jorder_e}, RESCAN\cite{rescan}) and proposed method (note that, we only chose to show the results of MENET-LB) on real rainy images.

Compared with synthesized images, rainy images from natural scenes are more complex. To present the comprehensive rain removal ability of all methods, we choose several real rainy images. Most of these images are kindly provided by previous de-raining work. According to different types of rain streak, we divide these real rainy images into four groups.

\textbf{Light rain (Drizzle).} 
In the first group, the shape of rain streak is thin. As shown in Fig. 12, we can notice that all methods are successful in removing the majority of rain streak except that JORDER-E tends to slightly over-smooth the background texture. And the proposed method deals with this case, as well as, if not better than, RESCAN. 

\textbf{Moderate rain.} When the speed of falling rain is slow, short rain streak can be captured by camera. The characteristics of those rain streak are short, bold and sparse. In Fig. 13, we compare all methods in clearing moderate rain streak, and our method and RESCAN get better performance. 

\textbf{Heavy rain.} Because of the rainy accumulation, it is hard to remove rain streak when rain is heavy, as shown in Fig. 14. Regardless of the interference with such accumulation, our method gets satisfactory results. 

\textbf{Strong light.} Rainy images sometimes are accompanied by strong light, which has a negative impact on the de-raining methods. In Fig. 15, we show the results processed by all methods. Despite that several rain streak in halation are too indiscernible to be detected, the proposed method gets satisfactory result. 

\textbf{Example of failure.} These methods progressively achieve better results in both quantitative and qualitative metrics. However, due to the limitation of their learning paradigm, namely training on synthetic rain and rain-free ground truth images, they might fail when dealing with some conditions of real rain streaks that have never been seen during training.

As above contrasts illustrate, it has been well supported that our method significantly competitive others and is successful in removing the majority of rain streak.  Because DSC only capture low level image features, they leave significant rain streak and rain artifacts in some rainy images. JORDER-E tends to result in drastically altered content and degraded texture, so that image details are lost or hardly visible. DDN uses high-level features to remove rain streak, but is limited by single task and smaller receptive field, so its results are not as good as ours. RESCAN achieves better results in most cases because of its effective use of context. It worthy note that the recommended method is no worse than RESCAN, if not much better than it. Namely, similar to RESCAN, our method significantly improves the visual effect of processed images also.

\section{Conclusion}

Rain removal in images is an important task in computer vision filed and attracting attentions of more and more people. The ultimate goal of this paper is to reduce the effect of rain streak so that the recovered image can be closer to the real images. Our basic model is followed by U-Net. We take advantage of multiple image characteristics and propose a multi-constraint framework for reconstruction of rainy images, especially for the images with heavy rain. We carefully design a multi-constraint loss function by incorporating the reconstruction loss, edge-aware loss and texture matching loss, at the same time designed two adaptive weighting algorithms to explore the further release of the model's potential. By minimizing the proposed multi-constraint loss function, the proposed MENET can obtain a more perceptually pleasing reconstruction with abundant textures and sharp edges. Although the recommended method does not have an absolute advantage in the evaluation indexes, the proposed method has visible improvement on perceptual result and SSIM/PSNR. And making the plausible images more reasonable and closer to ground truth.

\section*{}

\bibliographystyle{abbrv}
\bibliography{mybibfile}

\begin{thebibliography}{10}

\bibitem{multitask_learning}
R.~Caruana.
\newblock Multitask {Learning}.
\newblock {\em Mach. Learn.}, 28(1):41--75, July 1997.

\bibitem{he_instance_aware}
J.~Dai, K.~He, and J.~Sun.
\newblock Instance-aware semantic segmentation via multi-task network cascades.
\newblock pages 3150--3158, 2016.

\bibitem{hog}
N.~Dalal and B.~Triggs.
\newblock Histograms of oriented gradients for human detection.
\newblock 1:886--893, 2005.

\bibitem{ddn}
X.~Fu, J.~Huang, D.~Zeng, Y.~Huang, X.~Ding, and J.~Paisley.
\newblock Removing rain from single images via a deep detail network.
\newblock pages 3855--3863, 2017.

\bibitem{gatys2015texture}
L.~A. Gatys, A.~S. Ecker, and M.~Bethge.
\newblock Texture synthesis using convolutional neural networks.
\newblock {\em arXiv: Computer Vision and Pattern Recognition}, 2015.

\bibitem{gatys2016_style_transfer}
L.~A. Gatys, A.~S. Ecker, and M.~Bethge.
\newblock Image style transfer using convolutional neural networks.
\newblock pages 2414--2423, 2016.

\bibitem{fast_rcnn}
R.~Girshick.
\newblock Fast r-cnn.
\newblock pages 1440--1448, 2015.

\bibitem{he_guided_filter}
K.~He, J.~Sun, and X.~Tang.
\newblock Guided image filtering.
\newblock {\em IEEE Transactions on Pattern Analysis and Machine Intelligence},
  35(6):1397--1409, 2013.

\bibitem{he_resnet}
K.~He, X.~Zhang, S.~Ren, and J.~Sun.
\newblock Deep residual learning for image recognition.
\newblock {\em computer vision and pattern recognition}, pages 770--778, 2016.

\bibitem{he_identity_map}
K.~He, X.~Zhang, S.~Ren, and J.~Sun.
\newblock Identity mappings in deep residual networks.
\newblock {\em european conference on computer vision}, pages 630--645, 2016.

\bibitem{psnr}
Q.~Huynhthu and M.~Ghanbari.
\newblock Scope of validity of psnr in image/video quality assessment.
\newblock {\em Electronics Letters}, 44(13):800--801, 2008.

\bibitem{perceptual_loss}
J.~Johnson, A.~Alahi, and L.~Feifei.
\newblock Perceptual losses for real-time style transfer and super-resolution.
\newblock {\em european conference on computer vision}, pages 694--711, 2016.

\bibitem{kang_id}
L.~W. Kang, C.~W. Lin, and Y.~H. Fu.
\newblock Automatic single image based rain streaks removal via image
  decomposition.
\newblock {\em IEEE Transactions on Image Processing}, 21(4):1742--1755, Apr.
  2012.

\bibitem{nonlocal_means_filter}
J.-H. Kim, C.~Lee, J.-Y. Sim, and C.-S. Kim.
\newblock Single-image deraining using an adaptive nonlocal means filter.
\newblock pages 914--917, 2013.

\bibitem{non_local_derain}
G.~Li, X.~He, W.~Zhang, H.~Chang, L.~Dong, and L.~Lin.
\newblock Non-locally enhanced encoder-decoder network for single image
  de-raining.
\newblock {\em arXiv: Computer Vision and Pattern Recognition}, 2018.

\bibitem{scale_aware_derain}
R.~Li, L.~Cheong, and R.~T. Tan.
\newblock Single image deraining using scale-aware multi-stage recurrent
  network.
\newblock {\em arXiv: Computer Vision and Pattern Recognition}, 2017.

\bibitem{rescan}
X.~Li, J.~Wu, Z.~Lin, H.~Liu, and H.~Zha.
\newblock Recurrent squeeze-and-excitation context aggregation net for single
  image deraining.
\newblock pages 262--277, 2018.

\bibitem{lp}
Y.~Li, R.~T. Tan, X.~Guo, J.~Lu, and M.~S. Brown.
\newblock Rain streak removal using layer priors.
\newblock pages 2736--2744, 2016.

\bibitem{fpn}
T.~Lin, P.~Dollar, R.~Girshick, K.~He, B.~Hariharan, and S.~Belongie.
\newblock Feature pyramid networks for object detection.
\newblock pages 936--944, 2017.

\bibitem{fcn}
J.~Long, E.~Shelhamer, and T.~Darrell.
\newblock Fully convolutional networks for semantic segmentation.
\newblock {\em arXiv: Computer Vision and Pattern Recognition}, 2014.

\bibitem{dsc}
Y.~Luo, Y.~Xu, and H.~Ji.
\newblock Removing rain from a single image via discriminative sparse coding.
\newblock pages 3397--3405, 2015.

\bibitem{bsd_dataset}
D.~R. Martin, C.~C. Fowlkes, D.~Tal, and J.~Malik.
\newblock A database of human segmented natural images and its application to
  evaluating segmentation algorithms and measuring ecological statistics.
\newblock 2:416--423, 2001.

\bibitem{faster_rcnn}
S.~Ren, K.~He, R.~Girshick, and J.~Sun.
\newblock Faster r-cnn: towards real-time object detection with region proposal
  networks.
\newblock 2015:91--99, 2015.

\bibitem{unet}
O.~Ronneberger, P.~Fischer, and T.~Brox.
\newblock U-net: Convolutional networks for biomedical image segmentation.
\newblock pages 234--241, 2015.

\bibitem{subpixel_conv}
W.~Shi, J.~Caballero, F.~Huszar, J.~Totz, A.~P. Aitken, R.~Bishop, D.~Rueckert,
  and Z.~Wang.
\newblock Real-time single image and video super-resolution using an efficient
  sub-pixel convolutional neural network.
\newblock pages 1874--1883, 2016.

\bibitem{feqe}
T.~Vu, C.~Van~Nguyen, T.~X. Pham, T.~M. Luu, and C.~D. Yoo.
\newblock Fast and efficient image quality enhancement via desubpixel
  convolutional neural networks.
\newblock pages 243--259, 2018.

\bibitem{ssim}
Z.~Wang, A.~C. Bovik, H.~R. Sheikh, and E.~P. Simoncelli.
\newblock Image quality assessment: from error visibility to structural
  similarity.
\newblock {\em IEEE Transactions on Image Processing}, 13(4):600--612, 2004.

\bibitem{jorder}
W.~Yang, R.~T. Tan, J.~Feng, J.~Liu, Z.~Guo, and S.~Yan.
\newblock Deep joint rain detection and removal from a single image.
\newblock {\em computer vision and pattern recognition}, pages 1685--1694,
  2017.

\bibitem{jorder_e}
W.~Yang, R.~T. Tan, J.~Feng, J.~Liu, S.~Yan, and Z.~Guo.
\newblock Joint rain detection and removal from a single image with
  contextualized deep networks.
\newblock {\em IEEE Transactions on Pattern Analysis and Machine Intelligence},
  pages 1--1, 2019.

\bibitem{multi_task_face_recog}
X.~Yin and X.~Liu.
\newblock Multi-task convolutional neural network for pose-invariant face
  recognition.
\newblock {\em IEEE Transactions on Image Processing}, 27(2):964--975, 2018.

\bibitem{did-mdn}
H.~Zhang and V.~M. Patel.
\newblock Density-aware single image de-raining using a multi-stream dense
  network.
\newblock {\em computer vision and pattern recognition}, pages 695--704, 2018.

\bibitem{zhanghe_cgn_derain}
H.~Zhang, V.~A. Sindagi, and V.~M. Patel.
\newblock Image de-raining using a conditional generative adversarial network.
\newblock {\em arXiv: Computer Vision and Pattern Recognition}, 2017.

\end{thebibliography}

%
%


\end{document}